# From edges to meaning: Semantic line sketches as a cognitive scaffold for ancient pictograph invention

Author: Seowung Leem[1], Lin Gu[2], Ruogu Fang[1,3,4,5*]

[1]J. Crayton Pruitt Family Dept. of Biomedical Engineering, University of Florida, Gainesville, FL 32611, USA

[2] Research Institute of Electrical Communication, Tohoku University, Japan

[3] Dept. of Electrical and Computer Engineering, University of Florida, Gainesville, FL 32611, USA

[4] Dept. of Computer and Information Science and Engineering, University of Florida, Gainesville, FL. 32611, USA

[5] Center for Cognitive Aging and Memory, University of Florida, Gainesville, FL 32611, USA

## ABSTRACT

Humans readily recognize objects from sparse line drawings, a capacity that appears early in development and persists across cultures, suggesting neural rather than purely learned origins. Yet the computational mechanism by which the brain transforms high-level semantic knowledge into low-level visual symbols remains poorly understood. Here we propose that ancient pictographic writing emerged from the brain's intrinsic tendency to compress visual input into stable, boundary-based abstractions. We construct a biologically inspired digital twin of the visual hierarchy that encodes an image into low-level features, generates a contour sketch, and iteratively refines it through top-down feedback guided by semantic representations, mirroring the feedforward and recurrent architecture of the human visual cortex. The resulting symbols bear striking structural resemblance to early pictographs across culturally distant writing systems, including Egyptian hieroglyphs, Chinese oracle bone characters, and proto-cuneiform, and offer candidate interpretations for undeciphered scripts. Our findings support a neuro-computational origin of pictographic writing and establish a framework in which AI can recapitulate the cognitive processes by which humans first externalized perception into symbols.

## INTRODUCTION

Humans readily recognize objects even when they are reduced to sparse strokes or outlines. Infants and individuals with little prior knowledge can still identify abstract depictions, suggesting that our sensitivity to abstraction is not learned solely through culture, but is rooted in neural and evolutionary predispositions [1,2]. Neuroimaging studies further show that the visual cortex represents real scenes and abstract drawings in similar ways [3–5], and that regions supporting perception are also recruited during handwriting and drawing [4,6,7]. Yet, how different visual areas interact to transform perceptual input into visual production remains unclear.

Visual production can be viewed as a transformation from high-level semantic concepts to low-level visual codes. The hierarchical organization of the visual cortex, with feedforward encoding and feedback modulation, provides a biological substrate for this process: high-level areas convey semantic prior knowledge to lower areas, emphasizing essential boundaries while suppressing irrelevant details [8–12]. Converging evidence indicates that feedback from high area modulates gain and selectivity for low-level features as it reaches V1 or V2 [13,14]. Although computational vision models excel in feedforward perception tasks, the recurrent mechanism enabling humans to *simplify and generate* visual symbols has been less explored [15–17].

Throughout history, humans externalized perception into various forms of depiction [18–23] via drawings from cave paintings to pictographic scripts such as Egyptian hieroglyphs [24], Chinese oracle bone characters [25], and proto-cuneiform [26] (Figure 1). Despite cultural distance, early pictographs share structural similarity and often resemble modern perceptual sketches, implying a shared cognitive grounding rather than pure imagination. This raises a fundamental question: *Were pictographs shaped by neural constraints that favor simplified, boundary-based representations?*

Here, we hypothesize that ancient pictographs emerged from the brain's intrinsic ability to reduce visual stimuli into stable abstract forms. We build a biologically inspired digital twin of the visual hierarchy that first encodes an image into low-level features, generates a sketch, and iteratively refines it through feedback guided by semantic representations. The model produces symbols analogous to ancient pictographs and offers plausible interpretations for undeciphered scripts. Our work suggests a neuro-computational origin of pictographic writing and provides a framework in which AI can evolve alphabets the way humans once did.

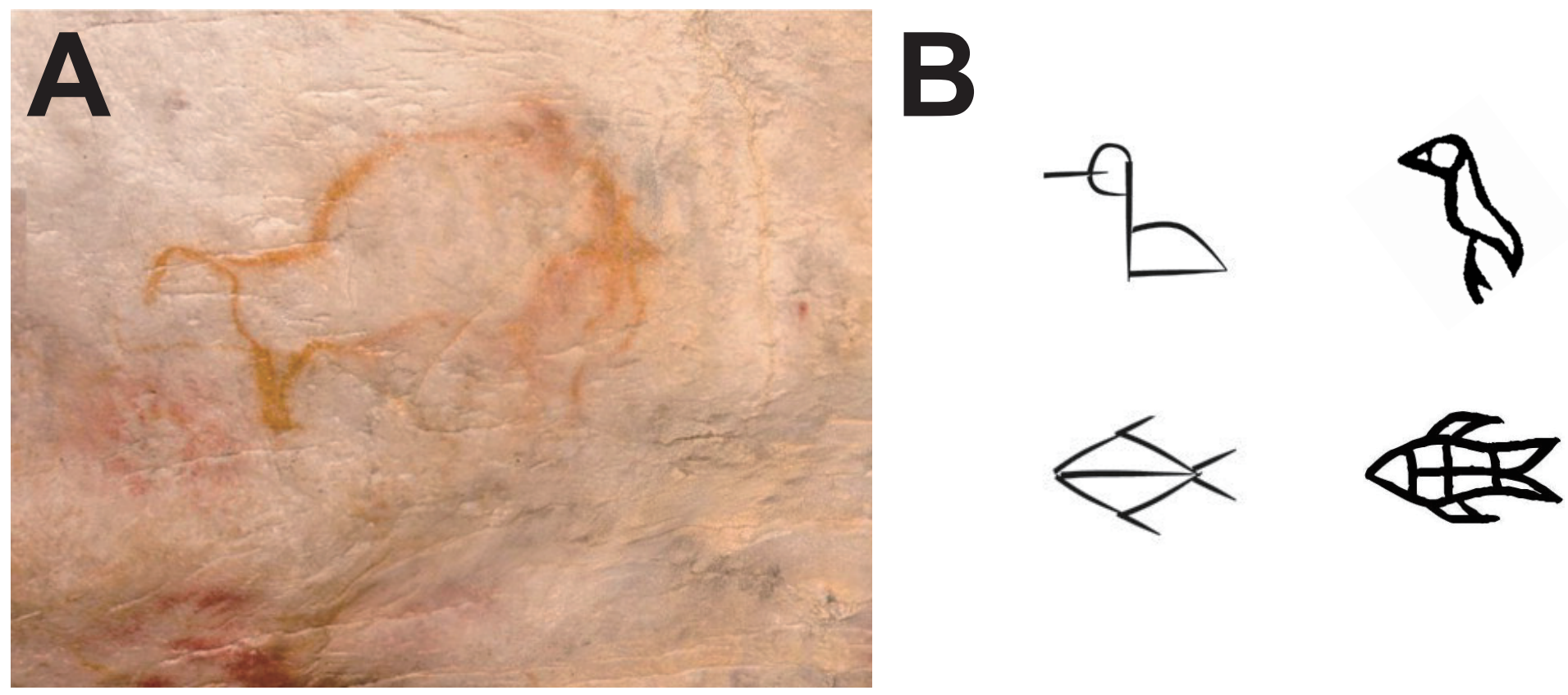

**Figure 1.** (A) Paleolithic cave painting of an animal found in a cave in Spain, Reproduced from Pike, A. W. G. et al. U-Series Dating of Paleolithic Art in 11 Caves in Spain. Science (2012) [18]. (B) Proto-Cuneiform (~3000 B.C.E) and Chinese Oracle (~12th B.C.E) of bird (Top) and fish (Bottom).

## METHODS

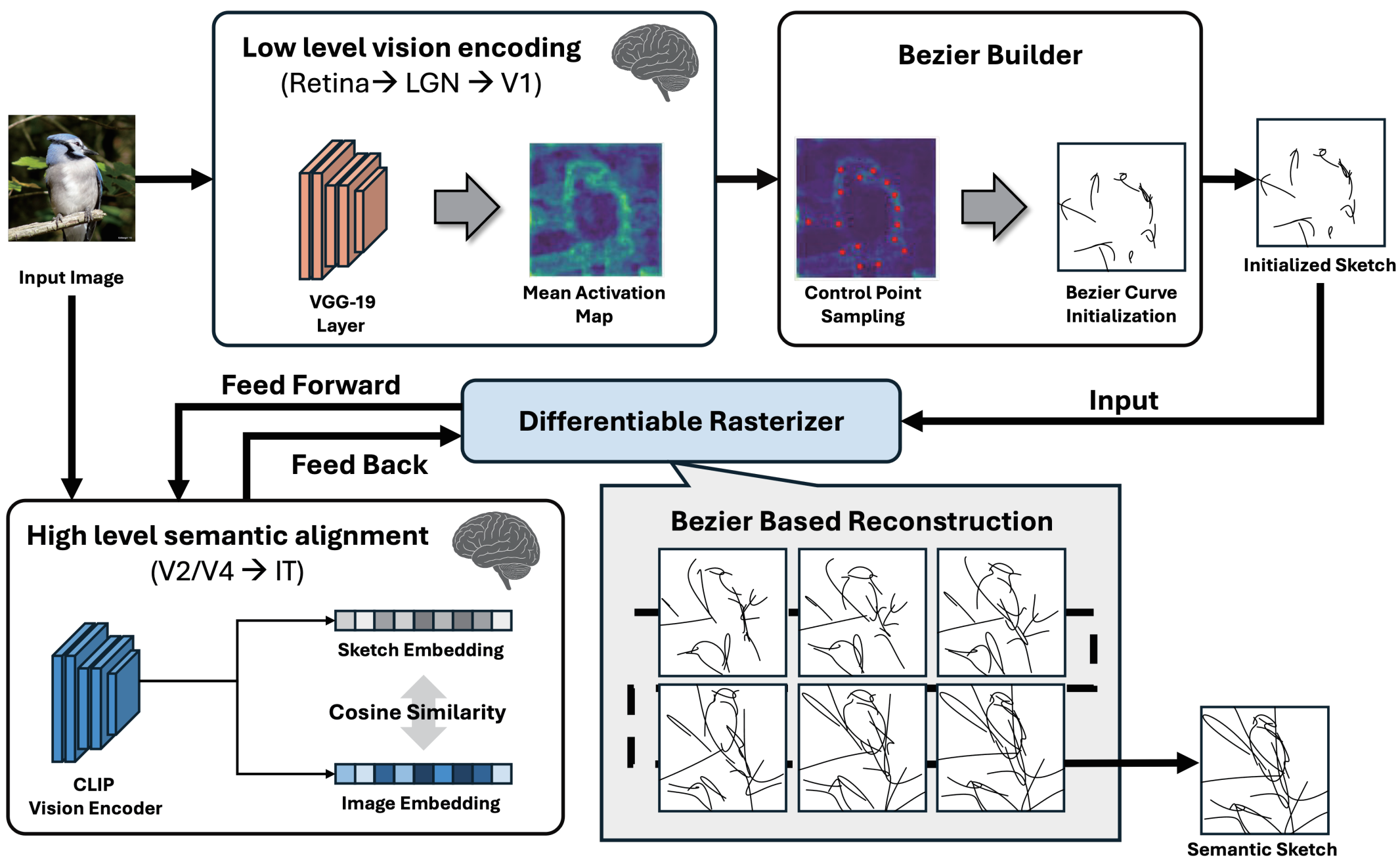

**Figure 2.** Digital twin framework for semantic sketch generation. The framework receives an image of an object or scene as input and outputs a semantic sketch represented as Bezier curve-based strokes. For low-level vision encoding, features from the first convolutional layer of VGG-19's third convolutional block were leveraged. From the convolutional block, the activation maps of the input were extracted and averaged to form a mean activation map. Bezier curve's start and control points were sampled from the activation map as a reference to initialize the sketch, designed to mirror the early visual response with respect to the input. A high-level vision module provides a semantic constraint to iteratively update the curve control points to generate a final sketch semantically aligned with the original input. The loss between the original input and generated sketch was computed by cosine similarity with embeddings from the vision encoder. The vision encoder from the vision-language model (CLIP) was used as a high-level vision model.

### Digital twin of human visual hierarchy

In this study, the digital twin model mirroring the visual production ability of the human visual hierarchy received the image $I \in R^{224\times224\times3}$ as an input and produced the Bezier curve-based sketch as an output (Fig. 2). The implementation of the digital twin was based on the vector graphic optimization framework by a differentiable rasterizer. [28]. Differentiable rasterizer allowed the vector parameter optimization using gradient-based techniques (e.g., backpropagation of loss) with specialized rendering of vectorized images into raster format. By leveraging the differentiable rasterizer, the cutting-edge raster-based techniques, such as CLIP, could be employed to edit the vector images. The objective of our digital twin was designed as a sketch generation process, semantically aligned with the input image. Our digital twin mimicked the low and high-level visual cortex with recurrent feedforward and feedback connections. The visual cortex comprised two distinct components: a low-level vision model initializing a sketch representing early visual activations given an input image, and a high-level vision model mimicking top-down semantic bias for optimization of the initialized sketch.

The low-level vision model was implemented by leveraging a convolutional neural network resembling the retina-to-V1 activation of primate visual cortex (Fig. S1). It was designed to derive the control point sampling process of Bezier curves to follow the salient regions of low-level visual activation, capturing the structural components of the input scene or object. The pipeline was built upon prior work on semantic sketching. [29]. The first layer of the third convolutional block from VGG-19 [30], the layer most resembling primate V1, was chosen as a target layer for activation map extraction. [31]. Given image I, activations from the 256-feature map were collected and averaged to acquire the mean activation map. The mean activation map was first min-max normalized and followed by a temperature (τ) controlled SoftMax function (τ = 0.3). The pixel with a higher SoftMax probability was visualized where the model expected informative information in the low-level visual context. The activation map was then leveraged to guide the sampling process of N strokes, each comprising k control points. Starting points of each stroke were sampled by a greedy algorithm with spatial suppression to prevent the regional clustering, points collapsing into a single hotspot. First, narrow image borders were excluded, followed by a sampling of the global maximum of the working probability map. After sampling, the normalized Gaussian kernel (σ = 5 pixels) at that location was subtracted before the next sampling process of the starting point. Following the starting point sampling, the remaining control points (k-1) were sampled by an attention-aware method to capture low-level visual structure. The square window with a square side length proportional to the image size (10% of the image size) was extracted from the attention map, centered on the starting point. From the center, local gradients were computed, followed by the tangent computation, which was formed as the vector perpendicular to the computed gradient. With the tangent, one pixel step was advanced following the tangent. Iteration proceeded until the trajectory reached the window boundary. When the point reached the boundary, the position was sampled as the next control point and updated as a new starting point to sample the remaining control points.

The vision encoder model trained from Contrastive Language-Image Pretraining (CLIP) was used as a high-level vision model (Φ) in our framework [32] (Fig. S2). CLIP jointly optimized the vision and text encoders with a contrastive objective in the shared embedding space, to align semantically similar image and text representations to be similar. Prior works had reported that the CLIP-based models performed better at capturing high-level brain regions. [33]. To semantically align the initialized sketch from the low-level vision model, each vector sketch was first rasterized in the RGB pixel space (S∈ R224×224×3). Then, the visual embeddings of both the original input and rasterized sketch were acquired by Φ. Then, the semantic loss (L semantic) was computed with Equation (1), which was the cosine similarity between the embeddings of the image and sketch. The computed loss was backpropagated to semantically align the sketch by updating the stroke parameters (control point positions). Update of the stroke parameters was performed 1500 times per sketch.

$$L_{semantic} = \frac{\Phi(\mathrm{I}) \cdot \Phi(\mathrm{S})}{\|\Phi(\mathrm{I})\|\|\Phi(\mathrm{S})\|} \tag{1}$$

**Dataset**

To validate our hypothesis, a standardized list of vocabulary shared across different regions was required. Thus, words of objects and scenes in the Swadesh list, a list used in lexicostatistics to compare the relationship between languages by identifying words sharing the same origin, were leveraged in our study. [34,35]. Lexical items from the Swadesh list were matched to object categories from the validation splits from the standard computer vision benchmarks (ImageNet, Microsoft COCO, Pascal VOC) [36–38]. This resulted in a stimulus corpus of 11 superordinate categories (bird, dog, fish, flower, fruit, lake, mountain, sand, sea, stone). The labels from the ImageNet, which were fine-grained, were collapsed into either of the superordinate categories. From Microsoft Coco and Pascal VOC, images from the dog and bird categories were retrieved (Table 1; full mapping information was provided in the Supplementary Materials). Retrieved images from the benchmarks were further stratified by how a human artist could capture the subject category and pose. For stratification, a pretrained vision transformer (ViT-B/32) classifier was finetuned with SketchyDatabase [39], consisting of 75,471 sketches with human ratings from 1 (very easy to sketch) to 5 (very difficult to sketch), with a classifier. The retrieved images from the benchmarks were stratified by sketchability. In this study, images with sketchability from 1 to 3 were used to generate the sketch.

| | **ImageNet** | **Microsoft COCO** | **Pascal VOC** | **Total** |
|---|---|---|---|---|
| **Bird** | 2950 | 192 | 131 | 3273 |
| **Dog** | 5900 | 556 | 428 | 6894 |
| **Fish** | 800 | - | - | 800 |
| **Flower** | 100 | - | - | 100 |
| **Fruit** | 500 | - | - | 500 |
| **Lake** | 50 | - | - | 50 |
| **Mountain** | 200 | - | - | 200 |
| **Sand** | 50 | - | - | 50 |
| **Sea** | 50 | - | - | 50 |
| **Snake** | 850 | - | - | 850 |
| **Stone** | 50 | - | - | 50 |

**Table 1. Source images by category from each computer vision benchmark.** Numbers in the table represent the images retrieved from the benchmarks (ImageNet, Microsoft Coco, and PASCAL VOC) for the study. The total in the right column is the sum of each category's images from the benchmarks. Dash indicates no image was collected from that source. The classes of objects (bird, dog, fish, flower, fruit, snake) and scenes (lake, mountain, sand sea, stone) were included.

| | **Sketchability: 1** | **Sketchability: 2** | **Sketchability: 3** | **Sketchability: Total** |
|---|---|---|---|---|
| **Bird** | 30 | 30 | 30 | 90 |
| **Dog** | 30 | 30 | 30 | 90 |
| **Fish** | 30 | 30 | 30 | 90 |
| **Flower** | 30 | 30 | 8 | 68 |
| **Fruit** | 30 | 30 | 30 | 90 |
| **Lake** | 9 | 30 | 3 | 42 |
| **Mountain** | 20 | 30 | 30 | 80 |
| **Sand** | 5 | 30 | 8 | 43 |
| **Sea** | 4 | 30 | 4 | 38 |
| **Snake** | 30 | 30 | 30 | 90 |
| **Stone** | 10 | 26 | 14 | 50 |
| **Category: Total** | 228 | 326 | 217 | 771 |

**Table 2. Images per category and sketchability.** The number of source images used for the sketch generation with three sketchability levels (1: easy to sketch~3: difficult to sketch). The right column gives a total number of images per category, and the bottom row shows the totals for each sketchability. Small counts under 30 reflect the limited availability of the corresponding category.

### Evaluation of semantic sketch generation

The evaluation of the framework was performed using qualitative and quantitative analysis. For qualitative analysis, sketches with different numbers of strokes (N = 4, 8, 16) across 11 object and scene categories were generated. For each category and stroke pair, up to 30 images were sampled. When the images were fewer than 30, all eligible images were used. The information on selected images was shown in Table 2. For qualitative analysis, the representative success and failure cases were reported. The sketch was labeled successful if a zero-shot classifier successfully ranked the ground truth category of the original input within its top three predictions, and failure otherwise Zero-shot classification was performed using a CLIP classifier with ViT-32/B as backbone, computing the distance between image embeddings and text embeddings with a text prompt defined as “A sketch of a(n) {class name}”.

The perceptual and semantic similarity between the original input and generated sketch pairs was analyzed by the representational similarity analysis (RSA) [40]. For the feature extractor, ImageNet Pretrained VGG-19 for perceptual representation, and the vision encoder from the CLIP model trained on large-scale image-text pairs for semantic representation were leveraged. The extracted features were normalized with the L2 norm. With normalized features, the cosine similarity matrix (S) was computed. The cosine similarity (s_(x,y)) between embedding x and y was computed by Equation 2, and mapped to the similarity matrix, followed by subtracting 1 to acquire the dissimilarity between feature representations. The perceptual and semantic similarity was quantified by Spearman’s rank correlation between the vectorized upper triangle without diagonal components between the original image and generated sketch RDMs. With the generated sketch and original image RDMs, ΔRDM was acquired by element-wise subtraction of the original image RDM from the generated sketch RDM. Additionally, the mean paired cosine similarity was computed to provide a summary of proximity between images and sketches in representational space. The statistical significance was acquired by a two-sided Monte-Carlo permutation test with 5000 permutations. For mean paired cosine similarity, the null hypothesis was acquired by breaking the pair of images and sketches. For RSA, a null distribution was acquired by randomly permuting the labels of the RDM from sketches.

$$s_{a,b} = \frac{\mathrm{x} \cdot \mathrm{y}}{\|x\| \|y\|} \qquad (2)$$

**Pictographs**

In this study, 2 ancient pictographs decoded in prior studies were compared with the generated sketch from the digital twin framework. The selection of the ancient writing system aimed at an analysis covering pictographs from different periods and regions. To meet this purpose, Egyptian hieroglyphs and Chinese oracle bone script (~12th B.C.E) were chosen [25]. For Hieroglyph, both the actual picture and the rendered image were acquired. Specifically, the picture of Hieroglyph extracted from the pyramid text found in the tomb of pharaoh Unas [41,42] and rendered text was acquired from a free, open-source word processor for ancient Egyptian hieroglyphic text (JSesh 7.9; https://jsesh.qenherkhopeshef.org/) [43]. For the Chinese oracle bone script, the open dataset for recognition and decipherment of oracle bone characters was leveraged [44]. Finally, the Proto-Cuneiform used in this study was sampled from the Cuneiform Digital Library Initiative (CDLI; https://cdli.earth/) [45,46]. The 2089 rendered proto-cuneiform image was acquired with a corresponding sign name. Then, the compound terms (n=662), numeric terms (n=343), and catalog terms (sign prefix with “ZATU”; n=242), which were not yet deciphered, were excluded, and 839 cuneiform signs were retained in this analysis.

| | Egyptian hieroglyphs | Count | Chinese oracle bone script | Count |
|---|---|---|---|---|
| **Bird** | G1 – G60 | 598 | 隹, 鳥 | 246 |
| **Dog** | E17 | 5 | 尨, 犬 | 153 |
| **Fish** | K1 – K10 | 10 | 魚 | 75 |
| **Flower** | M2, M8, M9, M11, M24A, M25, M26, M27, M28, M28A, M42, R20, R21 | 19 | 苑, 華, 芇, 辥 | 85 |

| **Fruit** | M39, M43 | 2 | 柚，栗 | 93 |
|---|---|---|---|---|
| **Lake** | N35, N35A, N37, N38, N39, N41, N42 | 442 | 泉，淵 | 93 |
| **Mountain** | N25, N26, N29 | 22 | 山，岳，嶽 | 228 |
| **Sand** | N18, N19 | 22 | 沙 | 24 |
| **Sea** | N35, N35A, N37, N38, N39, N41, N42 | 441 | 濤 | 28 |
| **Snake** | I9, I10, I11, I12, I14 | 167 | 它，巳，巴，虫 | 309 |
| **Stone** | W3 | 1 | 石 | 93 |

**Table 3. The codes of Egyptian hieroglyphs and Chinese oracle bone script for each category.** The code for the hieroglyph is based on the Gardiner's sign list. For oracle bone scripts, the modern Chinese character derived from the oracle bone character was denoted.

### Semantic matching and evaluation

To assess the similarity between the ancient pictographs and the generated semantic sketches, the pairwise similarity was computed in a shared visual-semantic feature space. Both pictographs and sketches were first encoded using the CLIP vision encoder model (ViT-32/B), which projected images into a common latent space. For sketches, the images retaining their ground-truth category within the top 3 zero-shot CLIP predictions were retained, ensuring that the embeddings reflected the reliable semantic cues. All pictographs were embedded without filtering. The zero-shot classification was performed with the same method as in the 'Evaluation of semantic sketch generation' section.

With embeddings, the pairwise matching was performed. Let $E_s \in R^{N_s \times d}$ denote the matrix of sketch embeddings and $E_a \in R^{N_a \times d}$ denote the semantic sketch embedding matrix, where $N_s$ and $N_a$ are the numbers of sketches and ancient pictographs, respectively, and $d$ denotes the embedding dimensions. The similarity matrix $M$ was computed as in Equation 3, where each entry $M_{ij}$ represented the semantic similarity between sketch $i$ and ancient pictograph $j$. Higher values indicated a stronger semantic correspondence between 2 different depictions.

$$M = E_s \cdot E_a^T \tag{3}$$

Followed by matrix computation, for each of the 11 categories, we identified the pictograph with the highest similarity to a given sketch. If the retrieved pictograph belonged to the same category as the query sketch, it was classified as a 'match' pair. Otherwise, the pair was labeled as a 'mismatch'. Matching accuracy was computed as the proportion of 'match' pairs across all sketches.

### Undeciphered Proto-Cuneiform arch sign sampling

To find the proto-cuneiform sign closely related to the generated sketches of 11 different categories, embedding-based sign sampling was performed. The same vision encoder from CLIP (VIT-32/B) was leveraged as an embedding model. For the embeddings of the generated sketch, a residual per-category embedding was computed. For each category, $c$ the embeddings of generated sketches $z_i$, were averaged by the number of sketches denoted by $n_c$ to compute the sum of normalized features ($v_c$) (Equation 4). This was followed by the normalization to unit length. Then, a global mean embedding ($d$) was computed across all categories as a baseline. To isolate the category's unique feature dimension, the linear revisualization was performed by subtracting the projection of each category's mean embedding onto the residual direction embedding to acquire the query embedding for each category ($r_c$) (Equation 5). The embeddings of proto-cuneiform signs ($c_j$) were simply acquired by passing the sign images into the encoder. Then, the cosine similarity between residualized query embedding ($r_c$) and cuneiform embeddings ($c_j$) was computed. Finally, the top 40 cuneiforms to query embeddings were acquired for each category.

$$v_c = \frac{1}{n_c}\sum_{i=1}^{n_c} z_i \tag{4}$$

$$r_c = v_c - \left(v_c \cdot \hat{d}\right), \hat{d} = \frac{d}{\|d\|} \tag{5}$$

## RESULTS

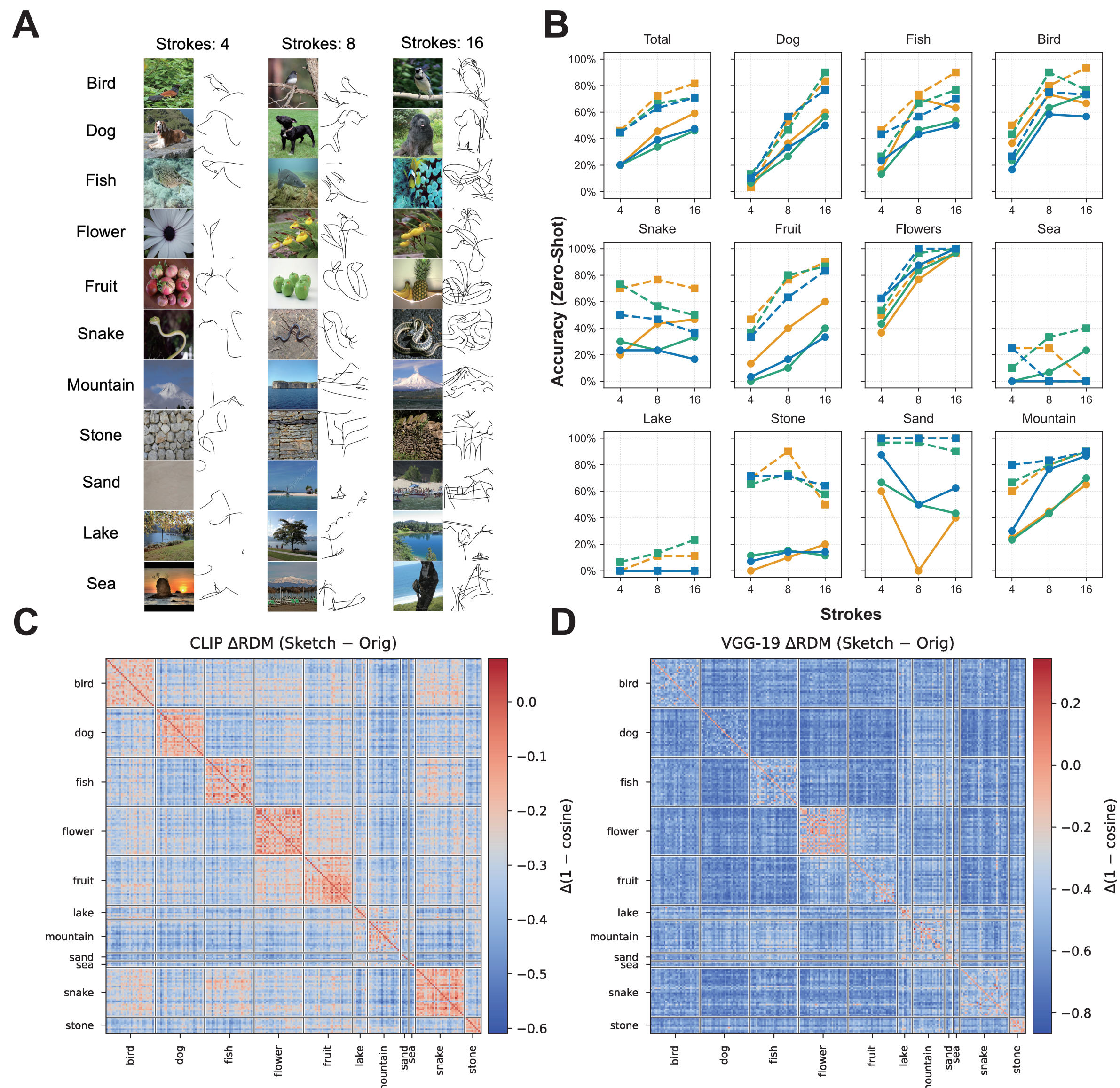

**Figure 3. Digital twin of the recurrent process of human visual hierarchy generates semantic sketches of input objects and scenes.** (A) The examples of input image and output sketch show the striking ability of the digital twin to generate a sketch resembling the input image with a limited number of strokes. Each row represents the categories used in our analysis, and each column shows the result with a different number of strokes. (B) Zero-shot classification accuracy of generated sketches across different sketchability and stroke counts indicates the different semantic characteristics of objects and scenes. The Top 1 (real line) and Top 3 accuracy across all and each category is presented. The different sketchability is denoted by colors (1: yellow, 2: blue, 3: green). Differences of representational dissimilarity matrices (ΔRDMs; higher values shown in red, denoting the high correspondence between sketch and image) from (C) CLIP (semantic) and (D) VGG-19 (perceptual) as an encoder indicate that the semantic representations better represent the categorical similarity. The value close to 0 (denoted as red) in ΔRDM indicates the high similarity between the sketch and the original input. The rows and columns represent each stimulus, and the black line represents the categorical boundary. Results are shown for sketchability level 1 with 16 stroke sketches.

**Semantic constraints drift the low-level strokes towards category-defining sketches.**

The digital twin framework was designed to receive the image of an object or scene as input and generate the semantic sketch as an output (Figs 2, S1, and S2). It consisted of two distinct modules, each mirroring the function of low-level and high-level human vision. The low-level vision module initialized a Bezier curve sketch based on the low-level visual activation patterns of the input scene. It was then recurrently updated by the high-level vision module with semantic constraints. The low-level vision was modeled by the convolutional neural network [30], explaining the V1 activation of primate visual cortex, and high-level vision was modeled by the vision encoder trained with Contrastive Language-Image Pretraining (CLIP) [32], resembling the high-level visual areas of human visual cortex [31,33]. The generated sketch was a Bezier Curve-based vector graphic, with black color, fixed stroke width, and control points (k) as 4 throughout this study. Based on the initialized sketch with a low-level vision activation pattern as the reference, the high-level semantic constraint leads the sketch generation process with a differentiable rasterizer [28]. Briefly, a differentiable rasterizer renders the vector image to raster format with differentiability, enabling the application of gradient-based methods to optimize the vector sketch parameters.

To generate the sketches, the list of categories in the Swadesh list [34,35] was leveraged, which is a list of forms and concepts that all language has terms for without exception. With a list, 11 categories from ancient Egyptian hieroglyphs [24] and Chinese oracle bone script [25] were selected. With categories, the corresponding images from the computer vision benchmark datasets were collected [36–38]. The collected images of each category underwent the sketchability labeling to determine the difficulty level (1: easy to sketch - 5: difficult to sketch) of capturing and drawing the target object from the image [39]. The image with sketchability from 1 to 3 was retained in this study. For each category, up to 30 images were sampled, and all eligible images were used if fewer than 30 were available. Further details of the framework and selection criteria for the natural image inputs are described in the method section.

Qualitatively, the model consistently reproduced the contours and structures of objects. The digital twin framework captured the semantic content of the input with a strikingly limited number of strokes. Specifically, with a minimal number of strokes (N=4), the generated sketches of birds, dog, fish, flowers, fruit, and snake retained the key structures (e.g., head and body separation, stem and leaf layout), resulting in recognizable sketches to both human and zero-shot classifiers. The increase in stroke numbers produced progressive refinements with additional edges augmenting semantic details without sacrificing overall context. The performance of the scene categories was variable. Mountains were often interpretable, but sketches of stone, sand, sea, and lake were often ambiguous. The scenes lacked a representative outline and often depended more on multiple object compositions. Thus, with limited strokes, the model failed to capture fine and distributed cues required for scene identity. Together, these observations indicate that the semantic constraints effectively steer the low-level visual representation towards semantically relevant visual production, particularly for distinct and contour-defined objects, while notably challenging the summarization of sparse and compositionally defined scenes.

Varying the number of strokes revealed a multi-dimensional organization of semantic information in generated sketches. In sketches with a minimal number of strokes (N=4), the coarse and diagnostic contours were preserved (e.g., bird's silhouette), sufficient to convey the identity information of the category without details. As the stroke number increased (N=8, 16), the model progressively introduced the detailed structure (e.g., wing, body, leg separation in birds; head and torso separation in dogs, petal and stem separations in flowers), indicating the model's priority in sketch generation. This progression was consistent across object categories (bird, dog, fish, flowers, fruit, snake). In scene sketches, mountains showed a similar trend with the number of strokes. Other scene sketches exhibited enrichment of compositions with an increase in strokes. As the number of strokes increased, the visual production process added elements, increasing the scene definitions (e.g., boats, palm trees, seabirds, coastline). This suggests the scene is encoded as a composition of constituent objects and their spatial relationships rather than a single outline.

Across sketches of superordinate categories with subcategories, sketches merged into a prototypical shape. In birds (Coucal, Junco, Jay; Fig. 3A), the generated sketch emphasized bird-level traits over the species-specific features. Fruits (Pomegranate, Granny Smith, Pineapple) converged toward an apple-like contour. An analogous trend was observed for dogs (Welsh springer spaniel, Staffordshire bull terrier, Newfoundland), fish (Puffer Sturgeon, Anemone), flowers (Daisy, Yellow lady's slipper), snakes (Vine snake, Thunder snake, Hognose snake), and mountains (Volcano and cliff). This convergence implies that the high-level semantic constraints drive the visual production to discard idiosyncratic details and preserve category-defining dimensions.

Recognition analysis mirrors the qualitative trends. Overall accuracy (zero-shot; Top 3 criterion) increases with stroke count (Total from Fig 3B; numerical values in Table S1). However, the optimal number of strokes differed by category. Object categories (dog, fish, fruit, flowers, mountain) reached their best performance with a greater number of strokes (N = 16), with a monotonic increasing trend from low to high numbers. Snake, sea, and lake exhibited only minor changes across different stroke counts, indicating that additional strokes did not yield sufficient category-diagnostic information. Stone and sand showed an inconsistent trend, reflecting their texture-dependent nature, which vector sketches capture poorly, as seen in the qualitative analysis. These results demonstrate that the semantic sketches generated by the digital twin were also interpretable by computational vision models, mirroring both the human capacity to recognize natural scenes and line drawings across formats, and the digital twin's ability to produce sketches legible to models with prior visual knowledge.

To assess whether sketch generation depends on the semantic constraints from the high-level visual areas, a control experiment was conducted in which the high-level vision module was replaced with the layer preceding the output layer of VGG-19. The embeddings of the input image and sketches were leveraged to provide a primarily perceptual constraint, guiding the framework to produce perceptual rather than semantic sketches. Qualitatively, the generated sketches for categories such as bird, flower, snake, and mountain exhibited comparable structures. However, recognition analysis results revealed that the top 1 and top 3 accuracies under semantic constraints consistently outperformed those from the perceptual condition (Fig. S3; numerical values in Tables S1, S2). Although certain scene categories (e.g., mountain, lake, sone, and sand) exhibited higher performance under the perceptual condition, these sketches lacked the refined structural and semantic coherence observed in the semantic results.

The perceptual and semantic correspondence between original images and generated sketch pairs was evaluated with feature representations from visual encoders with a learning scheme. By deriving representational dissimilarity matrices (RDMs) from a VGG-19 pretrained on the ImageNet dataset (perceptual encoder) and a CLIP vision encoder trained on OpenAI's large-scale image-text pairs (semantic encoder), representational similarity analysis (RSA) was conducted [40]. The difference between RDMs of sketch and image (ΔRDM) revealed that the semantic space better represents the categorical granularity. The generated sketch and original image have high semantic correspondence with robust categorical separation by exhibiting the vivid block structure along category boundaries (Fig. 3C). However, the perceptual correspondence between sketch and image was not aligned well (Fig. 3D), indicating the sketch has achieved the categorical separation despite low perceptual differences. This trend was coherent across all sketchability and stroke counts (Figs S4, S5). Moreover, the ΔRDMs derived from the semantic representation exhibited areas of high similarity regions (red regions) among the scene categories (e.g., lake, sand, sea), suggesting the greater representational overlap in the semantic space, due to ambiguity in the qualitative analysis of generated sketches.

Quantitatively, semantic similarity increased proportionally with the number of strokes (Fig. S6). The Spearman correlation between RDMs from original image and generated sketches (permutation test; $n=5000$) increased from low but significant values at 4 strokes (Sketchability: 1 = 0.091; $p<0.01$, Sketchability: 2 = 0.095; $p<0.001$, Sketchability: 3 = 0.108; $p<0.01$) to higher correlations at 8 strokes ($N=8$; Sketchability 1, $\rho = 0.260$; $p<0.001$, Sketchability 2, $\rho = 0.206$; $p<0.001$, Sketchability 3, $\rho = 0.280$; $p<0.001$) and 16 strokes ($N=16$; Sketchability 1, $\rho = 0.411$; $p<0.001$, Sketchability 2, $\rho = 0.355$; $p<0.001$, Sketchability 3, $\rho = 0.311$; $p<0.001$). In contrast, perceptual similarity was low at 4 strokes ($\rho < 0.021$; $p > 0.26$) and remained small at 8 (Sketchability 1, $\rho = 0.066$; $p<0.001$, Sketchability 2, $\rho = 0.005$; $p=0.802$, Sketchability 3) and 16 strokes (Sketchability 1, $\rho = 0.101$; $p<0.001$, Sketchability 2, $\rho = 0.061$; $p<0.01$, Sketchability 3, $\rho = 0.064$; $p<0.001$). The mean paired cosine distance between the original image-generated sketch pair mirrored this pattern (Fig. S6), implying that the generated sketches act as a semantic summary of objects and scenes, with a huge difference in perceptual structure.

**Cross-cultural evidence patterns of semantic grounding in pictographic writing.**

The digital twin of the human visual hierarchy demonstrated that top-down semantic constraints yield perceptually and representationally superior outputs relative to the original scene. To further support the hypothesis that semantic visual grounding underlies pictograph invention, we performed semantic matching between the generated sketches and ancient pictographs. While limited numbers of deciphered pictographs are currently available, the Egyptian hieroglyphs and Chinese oracle bone scripts were leveraged in this study. The pictures and rendered images of pictographs in the databases of previous studies were leveraged [41,43,44]. From the database, the pictographs of 11 different categories from the Swadesh list were acquired. For generated sketches, the samples with the top 3 criteria from zero-shot classification were retained. If the ancient pictographs were invented based on the semantic representational grounding, they should share similar semantics with the generated sketches, with the semantic constraints of their original scene. To validate this, the semantic matching between the generated sketches and the ancient pictograph was performed. Both sketches and pictographs were projected into the same embedding space, and the cosine similarity was used as a metric to measure the similarity between the sketches and pictographs. Then, if the generated sketch and the most similar pictograph fall into the same category, the pair was determined as 'match' and 'mismatch' if they failed to have the same category. The details of the ancient pictographs and the analysis method are described in the method section.

The semantic matching analysis revealed that several categories exhibited clear correspondence between generated sketches and ancient scripts, suggesting that their invention was guided by the semantic representational grounding. For both Egyptian hieroglyphs and Chinese oracles, at least 4 different categories reliably matched with generated sketches from the digital twin, respectively.

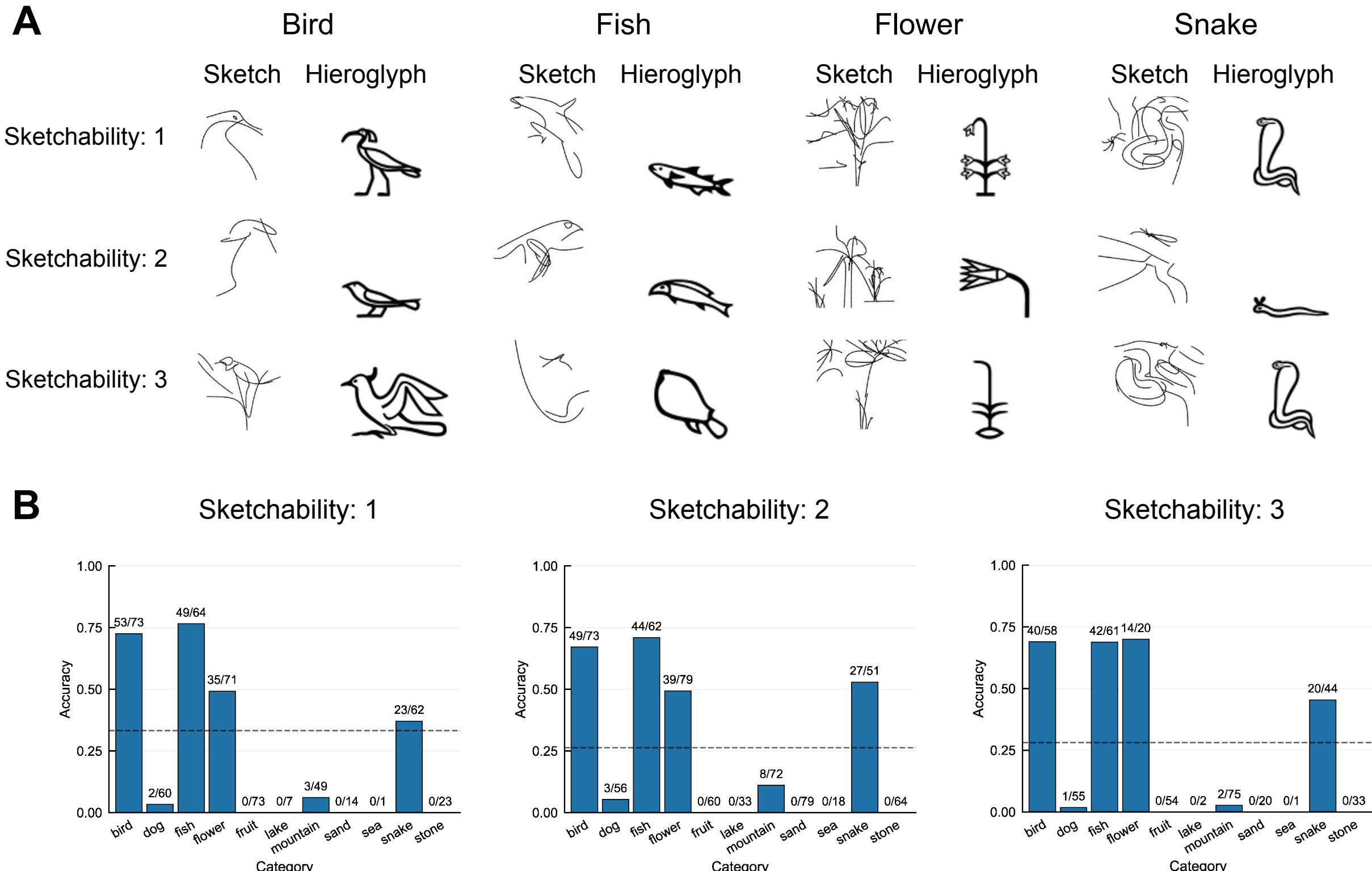

**Figure 4. Visual-semantic alignment between generated sketches and Egyptian hieroglyph pictographs.** (A) Representative examples of generated semantic sketches and their semantically aligned Egyptian hieroglyphs. Four categories are shown (bird, fish, flower, and snake), which exhibited the strongest cross-domain alignment. Within each category, examples of three sketchability levels are shown. Each column shows the pair of model-generated sketch (left) and its best-matched oracle character (right). (B) The matching pair accuracy across categories for each sketchability level. Each blue bar indicates the proportion of correct matches (correct pairs / total sketches for that category). The dotted line denotes the mean matching accuracy averaged across all categories.

Specifically, the pictographs of birds, fish, flowers, and snakes exhibited strong alignment with the generated semantic sketches, suggesting the visual form of these pictographs was grounded in the semantic representation of their referents (Fig. 4A). For instance, hieroglyphs of birds (matching pair; Sketchability 1: 53/73, Sketchability 2: 49/73, Sketchability 3: 40/58) and fish (Sketchability 1: 49/64, Sketchability 2: 44/62, Sketchability 3: 42/61) consistently ranked highest among the matches, reflecting the semantic representation of birds and fish that preserve the key morphological cues, such as beaks and wings for birds, and fins for fish. Likewise, the flower (Sketchability 1: 35/71, Sketchability 2: 39/62, Sketchability 3: 14/20) and snake (Sketchability 1: 23/62, Sketchability 2: 27/51, Sketchability 3: 20/44) hieroglyphs displayed high correspondence with generated sketches, supporting the idea that these characters originated from the direct visual abstraction of semantic representation. In contrast, categories such as dog (Sketchability 1: 2/50, Sketchability 2: 3/56, Sketchability 3: 1/55) and mountain (Sketchability 1: 3/49, Sketchability 2: 8/72, Sketchability 3: 2/75) represented limited correspondence with some individual exemplars matched semantically, but overall low similarity (Fig. 4B). This suggests that the generated sketch and pictograph invention went through different symbolic abstractions or depictions during the process. Finally, the categories including the fruit, lake sand, sea, and stone exhibited no matching pairs, implying that their pictographic forms were either highly abstracted or encoded in the joint of non-visual process.

The Chinese oracle had similar trends to Egyptian hieroglyphs. The categories, including bird (Sketchability 1: 44/73, Sketchability 2: 37/73, Sketchability 3: 26/58), fish (Sketchability 1: 11/64, Sketchability 2: 8/62, Sketchability 3: 5/61), fruit (Sketchability 1: 9/73, Sketchability 2: 15/60, Sketchability 3: 8/54), snake (Sketchability 1: 5/62, Sketchability 2: 6/51, Sketchability 3: 6/44), and stone (Sketchability 1: 6/23, Sketchability 2: 12/64, Sketchability 3: 7/33), showed a notable alignment, showing that the oracles closely resembled the corresponding sketches semantically. For instance, in the bird example, the matched sketch with oracles retained not only beaks but also identifiable wing configurations. Similarly, the fruit-related pictographs showed strong semantic overlap with preserving the features of fruit flesh and stalks (Fig. 5A). This shows the rendering of the semantic representation for both the digital twin and ancient humanity, preserving key attributes during the invention. In contrast, categories such as flower (Sketchability 1: 3/71, Sketchability 2: 5/79, Sketchability 3: 0/20) and mountain (Sketchability 1: 2/49, Sketchability 2: 4/72, Sketchability 3: 3/75) exhibited weaker but observable correspondence (Fig. 5B). Categories including dog, lake, sand, and sea displayed absent semantic similarity, implying that the invention of these characters underwent different derivation processes.

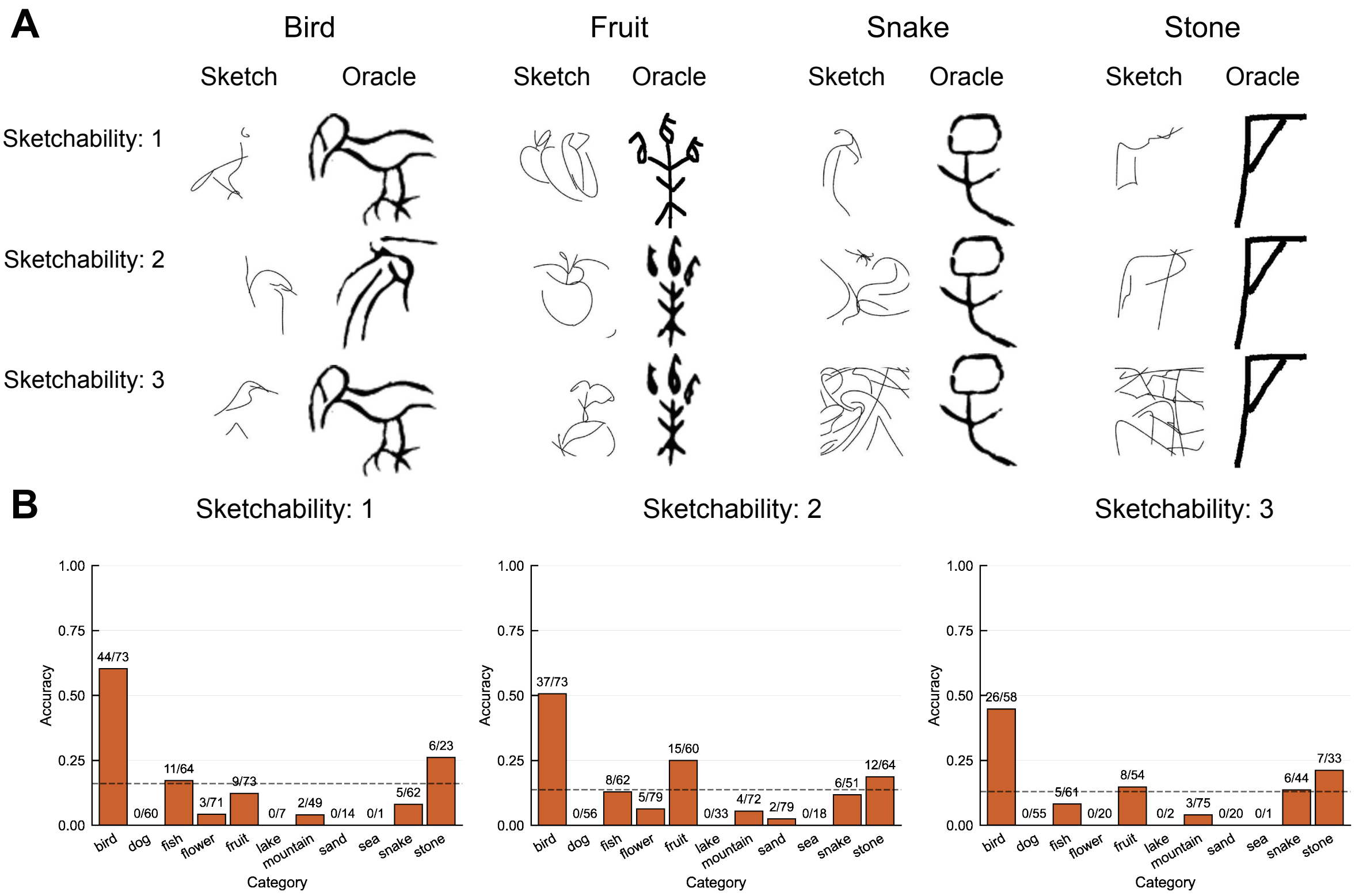

**Figure 5. Visual-semantic alignment between generated sketches and Chinese oracle bone pictographs.** (A) Representative examples of generated semantic sketches and their semantically aligned oracle bone characters. Four categories are shown (bird, fruit, snake, and stone), which exhibited the strongest cross-domain alignment. Within each category, examples of three sketchability levels are shown. Each column shows the pair of model-generated sketch (left) and its best-matched oracle character (right). (B) The matching pair accuracy across categories for each sketchability level. Each orange bar indicates the proportion of correct matches (correct pairs / total sketches for that category). The dotted line denotes the mean matching accuracy averaged across all categories.

**Semantic matching between proto-cuneiform scripts and sketch-based category features for decoding undeciphered pictographs**

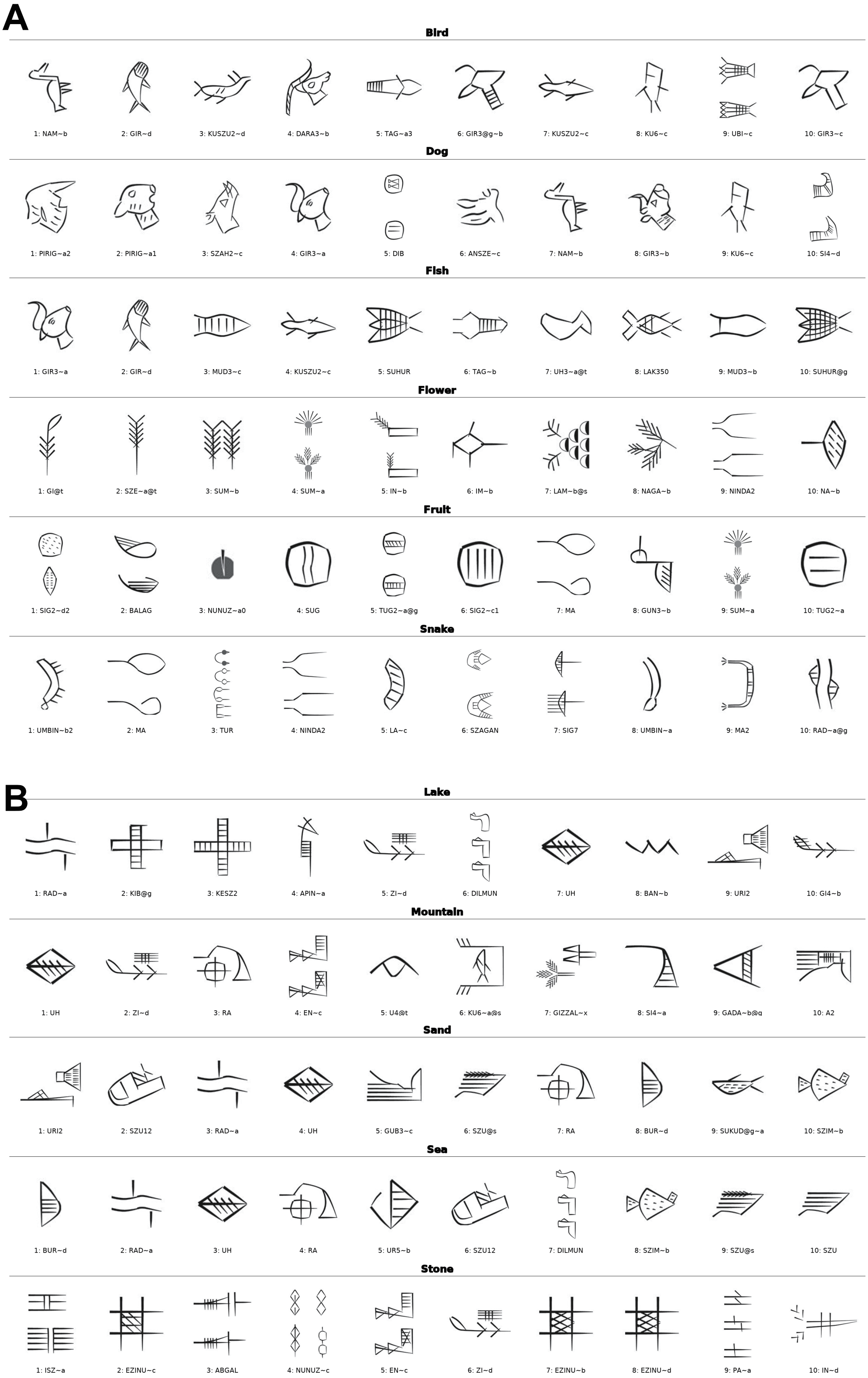

**Figure 6. Proto-cuneiform matches of each category were guided by the generated sketches.** Sampled Proto-cuneiform pictographs for object (A) and scene (B) categories were identified based on a residualized category derived from the generated sketch embeddings. The number below each pictograph indicates its rank, reflecting the degree of similarity to the residualized embeddings (smaller, more similar). The original sign names are annotated

beside each pictograph. The full top 40 pictographs are available in supplementary materials.

Finally, to evaluate whether the generated semantic sketches could be utilized in a decoding aid for undeciphered pictographs, the sketches from our framework were compared with the Proto-Cuneiform corpus. The generated sketches from 11 object and scene categories were transformed into the representative vector by residualized category embeddings computed from the mean sketch after regressing out the global average. These embeddings per category were compared with the embeddings of pictographs in a one-to-one manner. The top 40 pictographs (~top 5% from the pictographs used in the analysis) were extracted per category. This approach allowed us to assess the degree how much the semantic sketches corresponded to deciphered characters and to explore the potential of visual-semantic models for pictograph decoding.

The results revealed multiple cases of morphological or semantic convergence between generated sketches and Proto-Cuneiform pictographs (Fig. 6A; Full figure available in Figs. S7, S8). In the bird category, the most prominent example (Rank1: NAM~b; meaning "destiny; abstract marker") exhibited a strong visual resemblance to a bird sketch, while lower-ranked examples like GUN3 (Ranked 16; meaning "speckled, colorful") preserved avian characteristics despite a different meaning. In the dog category, the pictographs corresponded to the original meaning, like UR~c and UR~a (Ranks 15 and 26). In the fish, top matches included GIR~d (Rank 3), UBI~c (Rank 11), UBI~a (Rank 28), and SUHUR (Rank 10), all retaining a morphological characteristic of a fish. For flowers, the leading characters corresponded to either plant or crop lexemes such as GI@t (Rank 1: meaning "reed"), and SZE~a@t (Rank 2; meaning "barley"), while fruit included the produce marker NUNUZ~a0 (Rank 3; meaning "seed, egg"). In snake, the derivative pictographs such as BU~a (Rank 28; meaning "to tear out") and MUSZ3~a@g (Rank 35; meaning "surface, aspect") echoed the serpent form.

For scene categories, the sampling results captured broad contextual semantics compared to object categories (Fig. 6B). Pictographs associated with lakes and seas frequently exhibited hydric or geographic terms. For example, "DILUMN" (Rank 7 in Sea and Lake) describes the ancient kingdom of Dilmun emerged from the island inside the Persian Gulf region. Moreover, all seas, lakes, and mountains had mixed geographical contexts, including "KASKAL" (journey, road), "KALAM" (land). Stone category results displayed a grid or cross-hatched forms with their meaning mirrored by the administrative or status-related signs, including "EN" (Lord), "NIR" (Prince, Lord), "BARA2" (pedestal, dais, throne), reflecting the textural resemblance of stone surfaces rather than direct semantic overlap.

Across all categories, several high-frequency mismatches emerged. For instance, "SUKUD" (from bird, dog, sand, sea, and snake; meaning "tall; high"), and "RA" (from lake, mountain, sea, sand; meaning "to beat") emerged. However, these terms are adjectives that could diverge from the semantic groundings of corresponding categories. This indicates the framework's success in capturing graphical families, but it remained limited by the visual semantics. Nonetheless, semantically related pictographs emerged in meaningful contexts. For example, "SZANDANA~a", meaning gardener, appeared in the flower category (Rank 18), highlighting the semantic proximity. In dogs, large mammalian pictographs such as "UH3" (Ranks 18, 25), "SZEG9" (Rank 12), and "PIRIG" (Ranks 1, 2, 21) were retrieved, which were close to the canine's morphology. For the bird, "KAR2" (meaning "to steal; to escape") appeared in lower ranks (Rank 36), and in the snake, "UMBIN" (meaning "nail, claw, talon hoof") recurred across selected pictographs, suggesting the generalized semantics based on morphology or action features.

## DISCUSSION

This study provides the computational framework for the neuroscientific foundation of explaining how the invention of pictographs was driven by the visual semantic representation of the human brain. By modeling the recurrent dynamics of the visual hierarchical system, the proposed sketch generation framework offers a computational approach to discover neural mechanisms that are difficult to manipulate in vivo. In addition, this framework enables latent visual representations to be expressed in a perceptible and interpretable depiction, bridging human and computational perception without the need for additional decoding of complex features for understanding. This approach provides evidence that the semantic encoding of natural scenes in the human brain could serve as a cognitive mold driving the invention of pictographic forms. Notably, the generated sketches match closely with deciphered ancient pictographs from ancient Egypt and China, despite the model having no prior knowledge of the symbolic system or explicit depiction rules. The successful retrieval of semantically congruent pictographs with the mean semantic embeddings of the generated sketches underscores a shared representational structure between ancient pictographs and the model's semantic space aligned with high-level human visual processing. This convergence suggests the semantic mappings captured by the model may reflect the cognitive principles that guided the formulation of the early written symbols.

This study extends the prior evidence demonstrating the shared representational geometry of the visual and linguistic information within the human visual cortex [47,48]. Visual object representation integrates both spatial and semantic information [49,50], hierarchically built on top of the contextual information of the object [51]. Decades of research on the visual pathway have revealed that the higher-order cortical regions encode objects along the continuum from low-level perceptual features to abstract semantic attributes [52–54]. In line with these findings, the framework exhibits progressively enriched representational detail as the number of strokes in the framework increases (4, 8, and 16). This indicates the expressiveness of the semantic content scales with the model's representational capacity, defined by the number of strokes in this study. This observation reinforces the notion that semantic representations are distributed across multiple hierarchical levels, bridging perceptual structure and conceptual meaning.

The poor quantitative performance for scene sketches arose because the framework tended to render constituent objects rather than the scene as a whole (e.g., a body-of-water scene was rendered with a parasol, palm trees, and a swimming fish). This implies that scene representation is fundamentally linked to object representation, consistent with the existence of separate cortical pathways for scene and object perception[55,56]. Previous studies have shown that object representations are embedded with contextual information of other objects for better visual functions in the ambiguous and challenging tasks [57,58]. The generated sketch of the scene categories from the study (sea, lake, and sand) provides additional support that builds on the previous study. Additional research into the process of scene perception in the human visual cortex should be considered, given the existence of the distinct brain areas dedicated to this function [56].

The matching results of the generated sketch and ancient pictographs align with the evolutionary perspective on how humans have acquired the specialized brain regions for reading. The visual word form area (VWFA), which lies in the ventral visual pathway, acquired the capability by re-establishing the selectivity of neurons previously tuned for object scenes or faces [59]. According to the neuron recycling theory, this repurposing process of object-selective neurons was induced by the domain specialization capabilities (e.g., reading), which was too recent to have specialized brain region [60]. This recycling process could have been initiated by pictograph invention and subsequently elaborated into the VWFA as pictographs evolved into modern writing. Our findings thus offer a neuro-computational account of the origins of human literacy by connecting semantic visual representation with ancient pictographic systems.

The strong semantic alignment between the generated sketches and ancient pictographs highlights the potential of this framework as a computational tool for assisting the decipherment of writing systems whose meaning has yet to be decoded. While scripts such as Egyptian hieroglyphs and Chinese oracle bone scripts have been extensively studied, others, such as scripts from the ancient Indus [61] and Elamite [62] civilizations, remain partially or entirely undeciphered. Traditional decipherment systems begin with the identification of the closest visual analog, followed by inferring the meaning through archaeological discoveries or cultural and historical semantics [63–65]. By quantifying distances in the latent representational space between ancient pictographs and model-generated sketches, our framework can substantially accelerate this process by efficiently narrowing the search for visual correspondence. Crucially, grounding the search in semantic space offers a richer basis for visual correspondence than purely perceptual features alone. This approach thus introduces a novel computational pathway for probing the semantic foundations of early symbolic communication systems.

Altogether, this study advances the hypothesis that the origins of ancient pictographs are grounded in the visual-semantic representation of the human visual cortex. Through qualitative and quantitative analysis across three distinct pictographs, the study offers a novel interdisciplinary perspective linking the evolution of symbolic communication with neuroscientific principles. Future work should aim to rigorously validate this hypothesis through precise computational modeling and empirical analyses of the visual cortex. Beyond illuminating the cognitive roots of pictograph invention, this study exemplifies how artificial intelligence (AI) can be leveraged as a digital twin of the brain, providing a powerful platform for testing the neuroscientific basis of any ancient inventions that were considered serendipitous. This fresh direction stands to provide new insights in both computer science and archaeology by revealing how representational principles in the human brain translate to cultural and technological innovations.

## LIMITATIONS OF THE STUDY

This study proposes that visual semantic grounding is a neural basis of ancient character invention. Although our findings are striking, there is a need for additional work to address the limitations of this study. Ancient characters have different roots based on their creation method. While pictographs are designed to resemble the object they originated from, there are ideographic (concepts), phonographic (sounds), and compounds that emerged in the ancient symbolic system. One of the failure examples of sketch generation and semantic matching of scene categories was primarily because pictographs such as sea and lake were composed of water and land. Moreover, there are more than visual semantics that influenced the ancient pictograph generation. Across object or scene categories of proto-cuneiform analysis, multiple incidents of unknown pictographs emerged in the sampled results. For instance, “ENZINU” appears five times (Stone, Mountain); “GIR3” appears six times (Bird, Dog, Fish) “KUSZU2” shows seven times (Bird, Fish, Sand, Snake), “RAD” occurs six times (Lake, Mountain, Sand, Sea), and “SIG2” turns up ten times (Dog, Fruit, Sand). These recurrence patterns suggested latent semantic grouping by related objects and scenes, but the current framework is limited by the visual semantics to fully explore the additional semantics for full decipherment.

Another limitation is the lack of biological realism of the framework used in this study. The digital twin designed in this study is primarily focused on the functional modeling of the visual cortex with a simplified visual hierarchy. In primates and humans, the visual processing spans interconnection between extensive feedforward, lateral, and feedback connections [8,10,66–68,] limiting the biological fidelity of the framework. In addition, despite the AI model’s comparable performance and representational correspondence with human visual representations, the explainability of the model is limited [69]. Since the framework has leveraged the models' success in perceptual representation, the influence of contextual processing, attentional modulation, and generalization to challenging scenes in pictograph invention requires additional analysis. In terms of biological realism, the simplified version of neural reality designed in this study limits the interpretation of the result from a neurophysiological perspective. The update rules using backpropagation are unlikely to be the learning rules of the human brain and lack the neuroanatomical constraints [70].

Finally, the resemblance found in this study does not imply that our model re-creates the historical invention process. The drawing and writing involve a complex sensorimotor integration with vision [71]. The current framework focuses on the computational modelling of the visual hierarchy and is limited in its account of how motor function influences the visual cortex during visual production. Moreover, our framework does not incorporate cultural influences on writing system invention; therefore, the interpretation and significance of our findings should be approached carefully [72]. Therefore, although the result provides an intriguing clue to neural principles, the framework does not provide a complete explanation of how pictographs arose in the ancient civilisations. Future work should address these limitations to more fully capture the neural and cognitive processes underlying visual symbol creation.

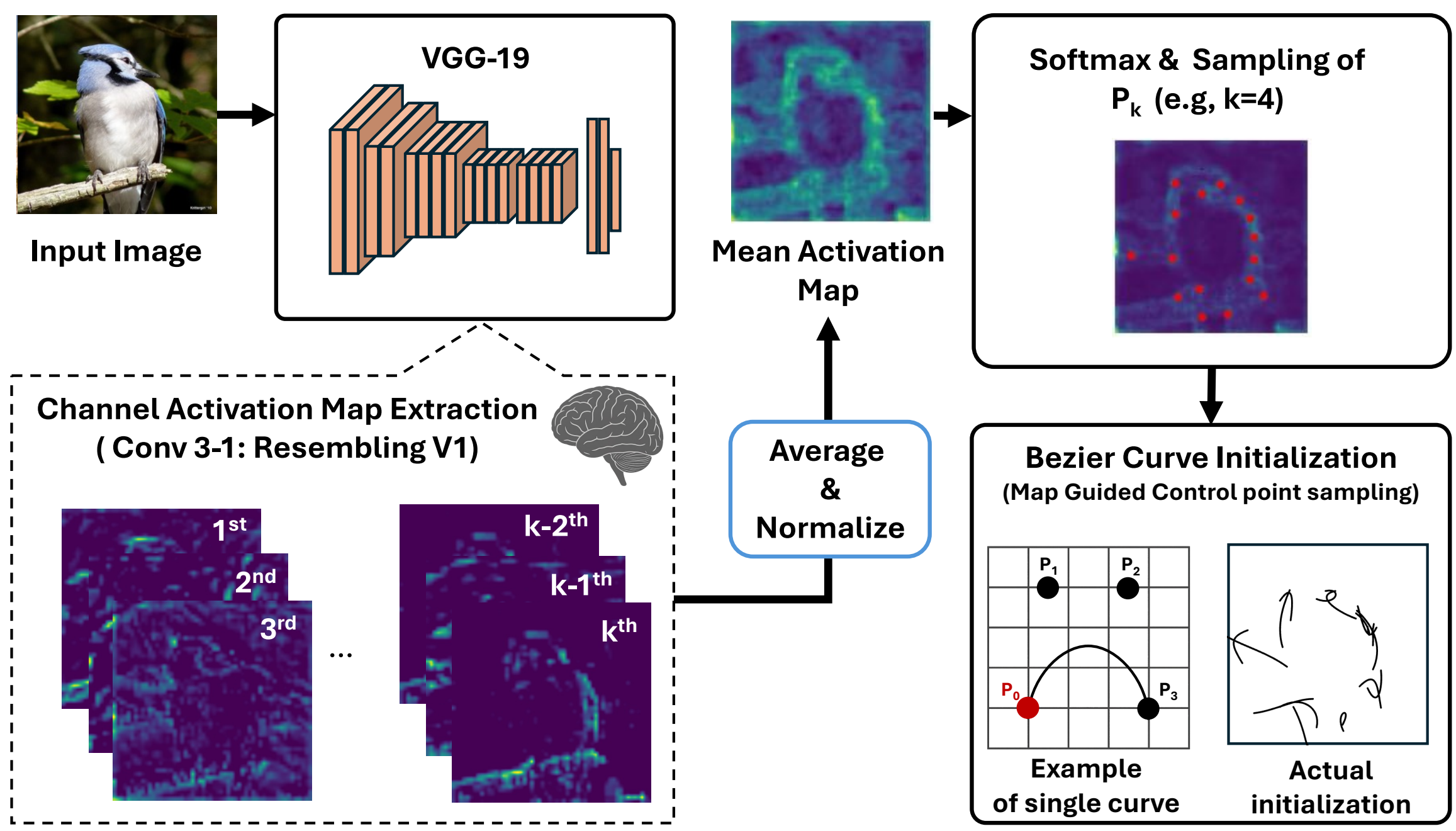

**Figure S1. The details of low-level vision module for Bezier curve based sketch initialization resembling low-level human vision (Related to Figure 2)**. From VGG-19, the first convolutional layers from the third convolutional block were leveraged as a feature extractor to acquire activation map for each channel (k=256). Then, the activation map was normalized and averaged to acquire the mean activation map which was designed to resemble the low-level human visual perception. Then, the SoftMax function was applied to the feature map, followed by the sampling of control points for Bezier curve initialization. The red points denote the start point of each Bezier curve ($P_0$; starting point example of 16 strokes sketch initialization.)

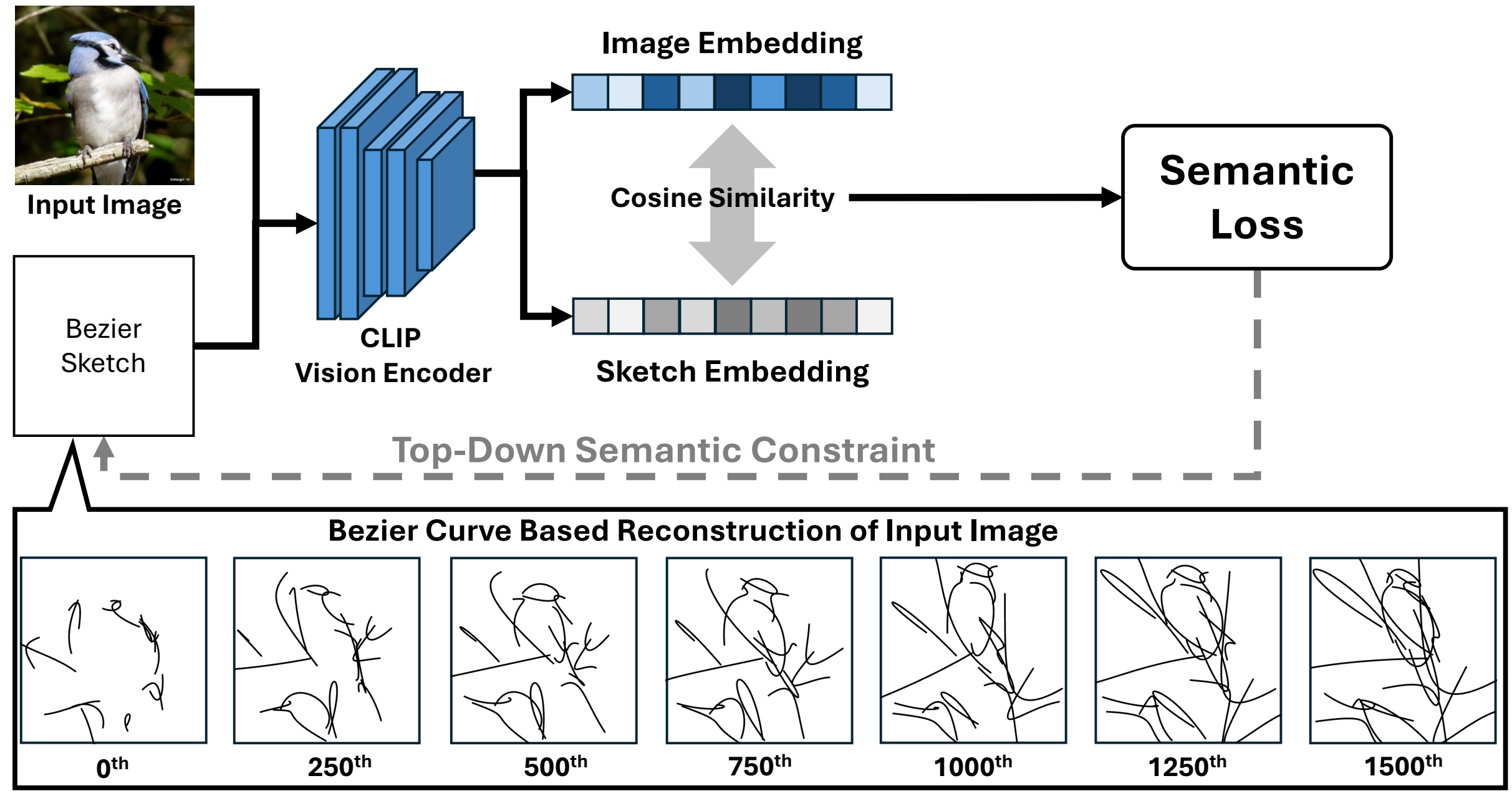

**Figure S2. The details of high-level vision module for Bezier sketch updates based on the semantic constraint (Related to Figure 2)**. The vision encoder from contrastive language image pretraining (CLIP) was leveraged, providing the top-down semantic constraints for Bezier sketch updates, designed to model the recurrent processing of high-level human vision. The constraint is provided in the format of loss function defined by cosine distance, and the objective of the learning was defined by decreasing the cosine similarity between the semantic embeddings of image and sketch. The example of 0, 250, 500, 750, 100, 1250, and 1500$^{th}$ of a bird sketch with 16 strokes were shown here for the example.

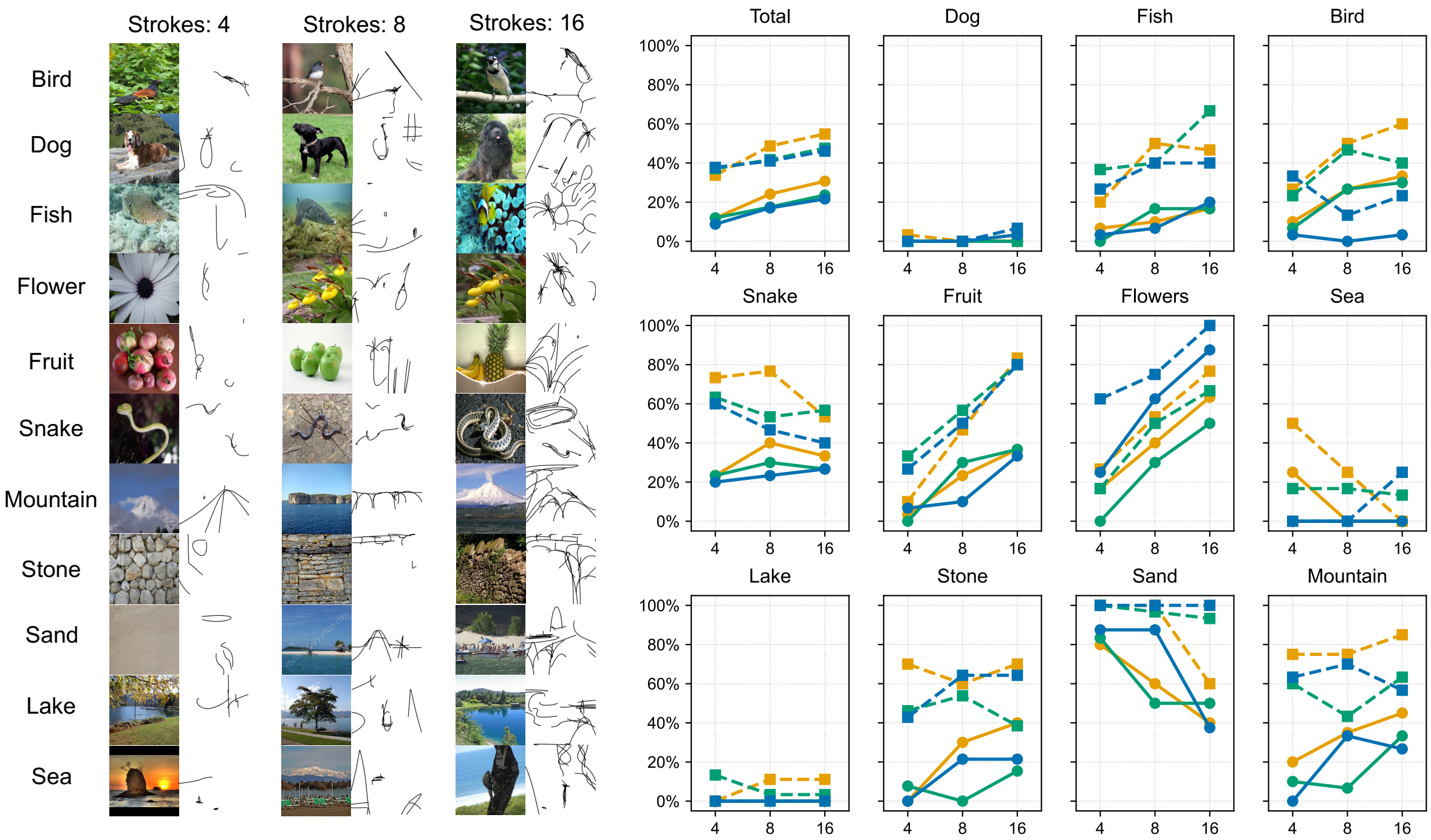

**Figure S3. The perceptual sketch generation result and zero-hot recognition result (related to Figure 3A and 3B).** (A) The examples of input image and output sketch show the perceptual constraint is not the neural basis of the visual abstraction of the input scene for pictographic formation. (B) Zero-shot classification accuracy of generated sketches across different sketchability and stroke counts indicates the perceptual sketches underperforms compared to semantic sketches. The Top 1 (real line) and Top 3 accuracy across all and each category is presented. The different sketchability is denoted by colors (1: yellow, 2: blue, 3: green).

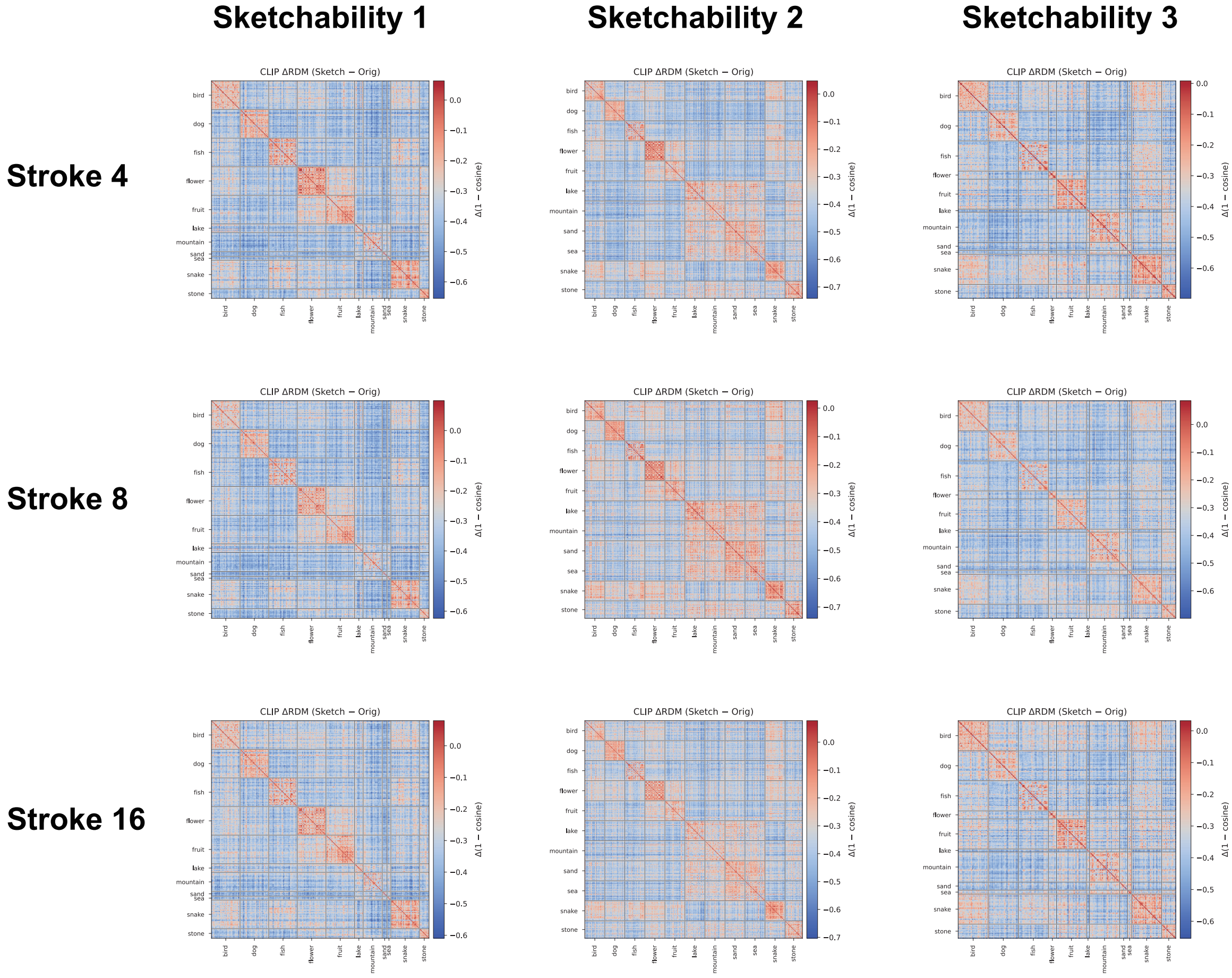

**Figure S4. Category-wise representational correspondence in semantic space across stroke counts and sketchability (Related to Figure 3C).** ΔRDMs depict the representatioal difference between the sketch and original images are shown based on the sketchability (columns) and stroke counts (rows). Each matrix shows the correlation difference between CLIP vision encoder embeddings, where bigger values (red) indicate the high similarity between the sketch and original representations. Category boundaries are denoted as black grid lines.

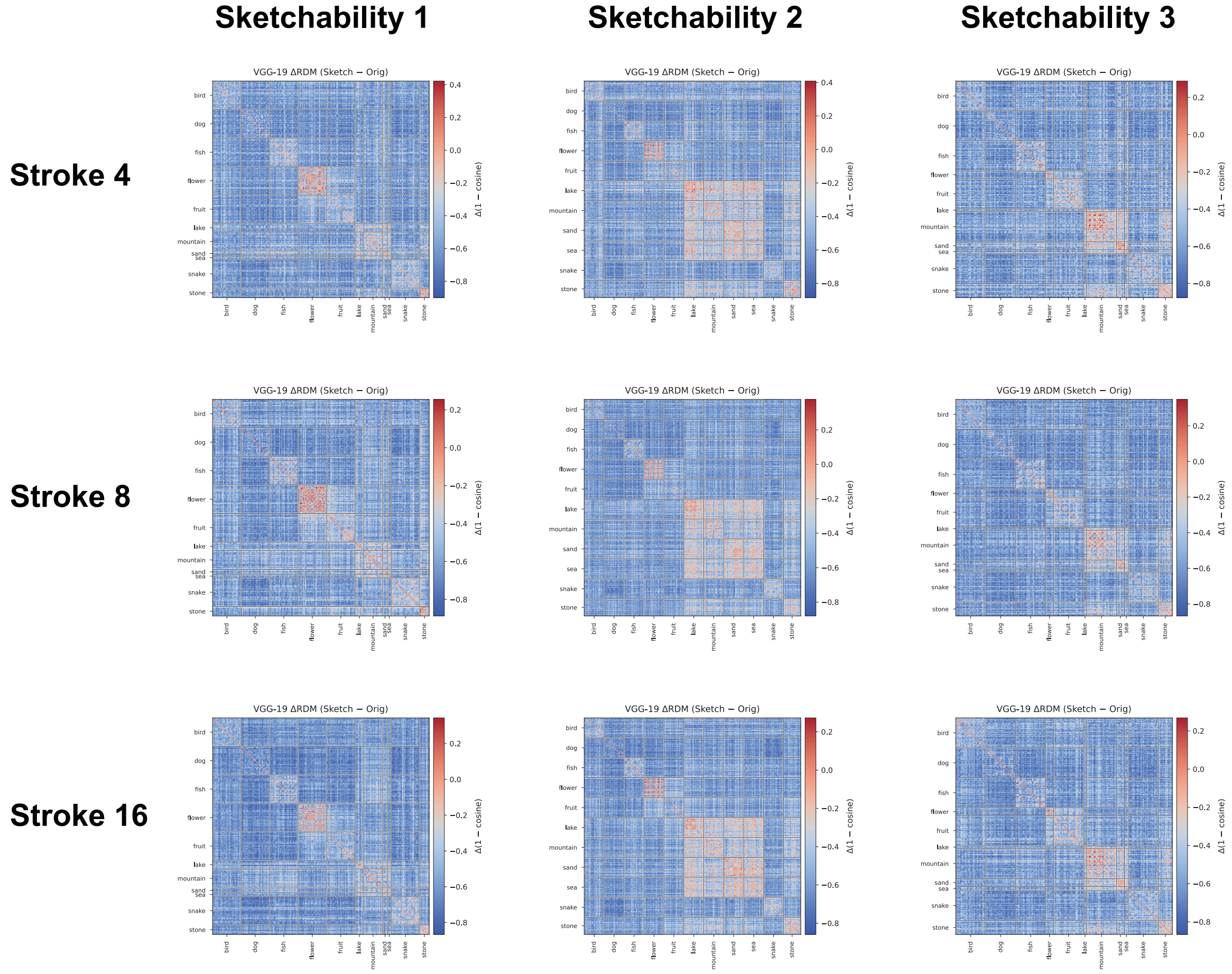

**Figure S5. Category-wise representational correspondence in perceptual space across stroke counts and sketchability (Related to Figure 3D).** ΔRDMs depict the representatioal difference between the sketch and original images are shown based on the sketchability (columns) and stroke counts (rows). Each matrix shows the correlation difference between VGG-19 encoded embeddings, where bigger values (red) indicate the high similarity between the sketch and original representations. Category boundaries are denoted as black grid lines.

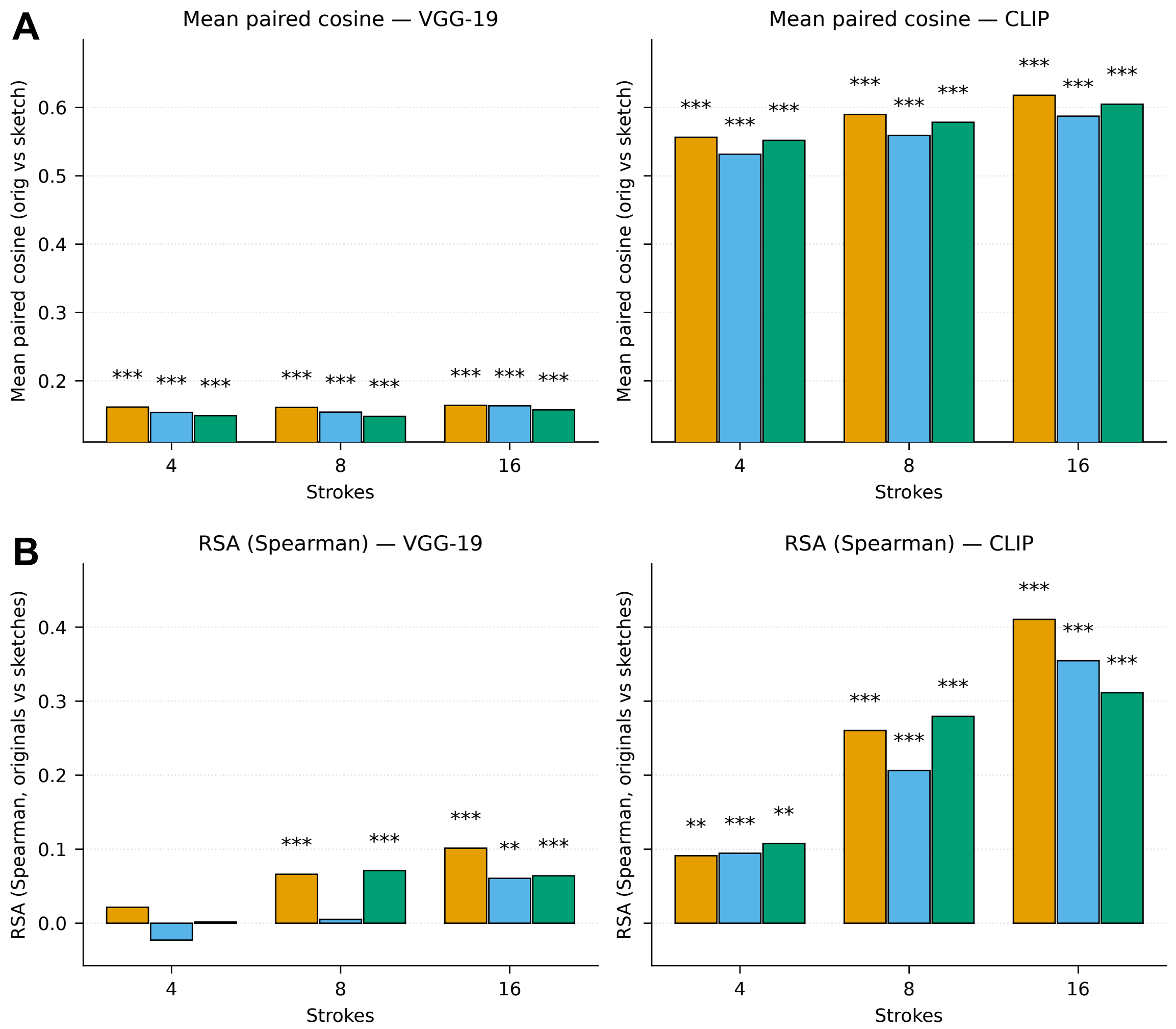

**Figure S6. Qualitative analysis of representational alignment between generated sketch and original image.** (A) Mean paired cosine similarity between the original image and corresponding generated sketch across stroke counts (4, 8, 16) is shown for perceptual (left; VGG-19) and semantic (right; CLIP) spaces. (B) Representational similiarity analysis (RSA) comparing the representational dissimilarity matrices (RDMs) of original image and generated sketch is computed using Spearman's rank correlation. The colors denote the sketchability levels (1: yellow, 2: blue, 3: green). Higher values indicate stronger representational alignment. Statistical significance is denoted via significance threshold (** $p < 0.01$, *** $p < 0.001$) acquired by permutation test (n=5,000).

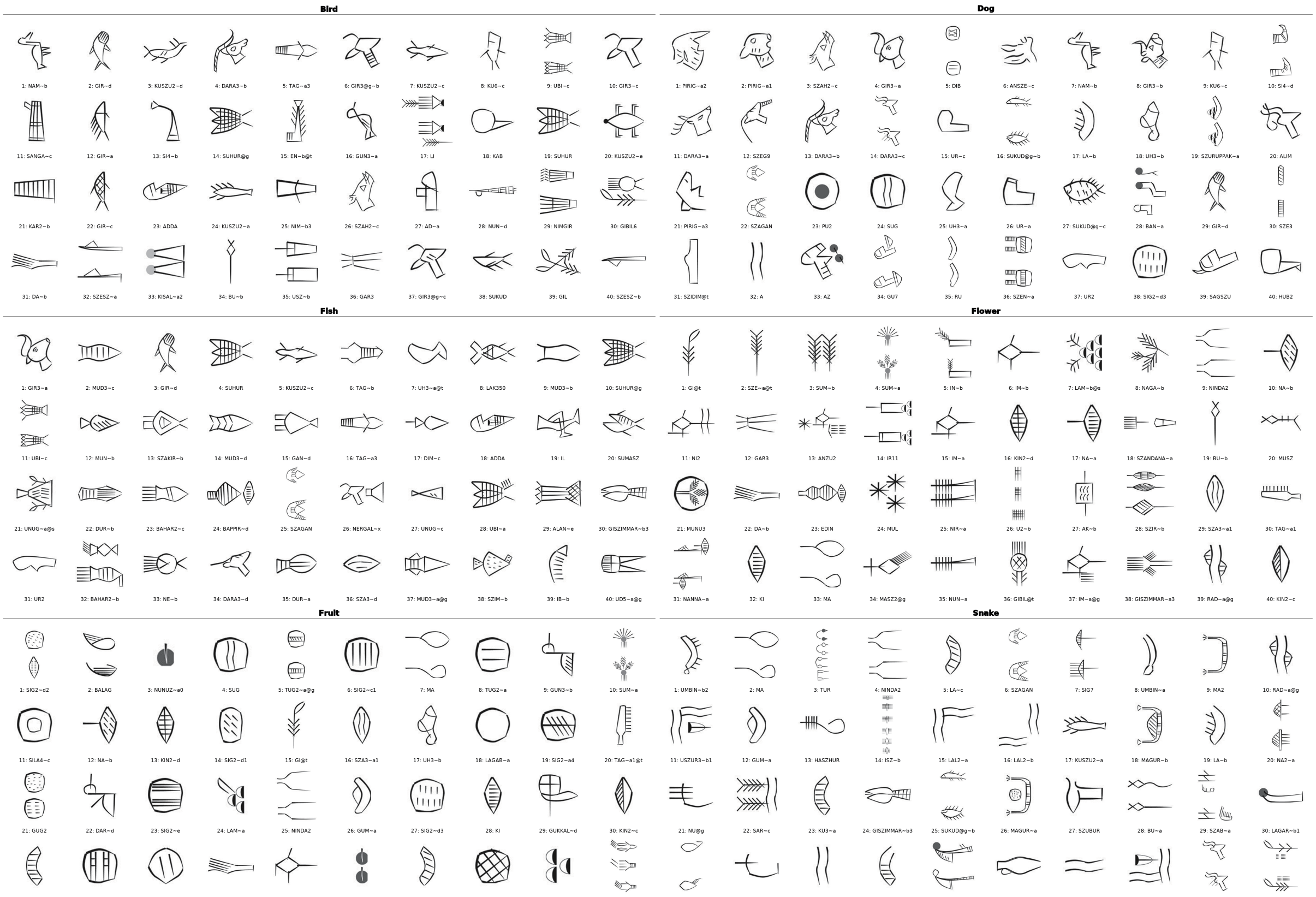

**Figure S7. Top 40 proto-cuneiform matches of object categories guided by the generated sketches.** Sampled Proto-cuneiform pictographs for object categories identified based on residualized category derived from the generated sketch embeddings. The number below each pictograph indicates its rank, reflecting the degree of similarity to the residualized embeddings (smaller, more similar). The original sign names are annotated beside each pictograph.

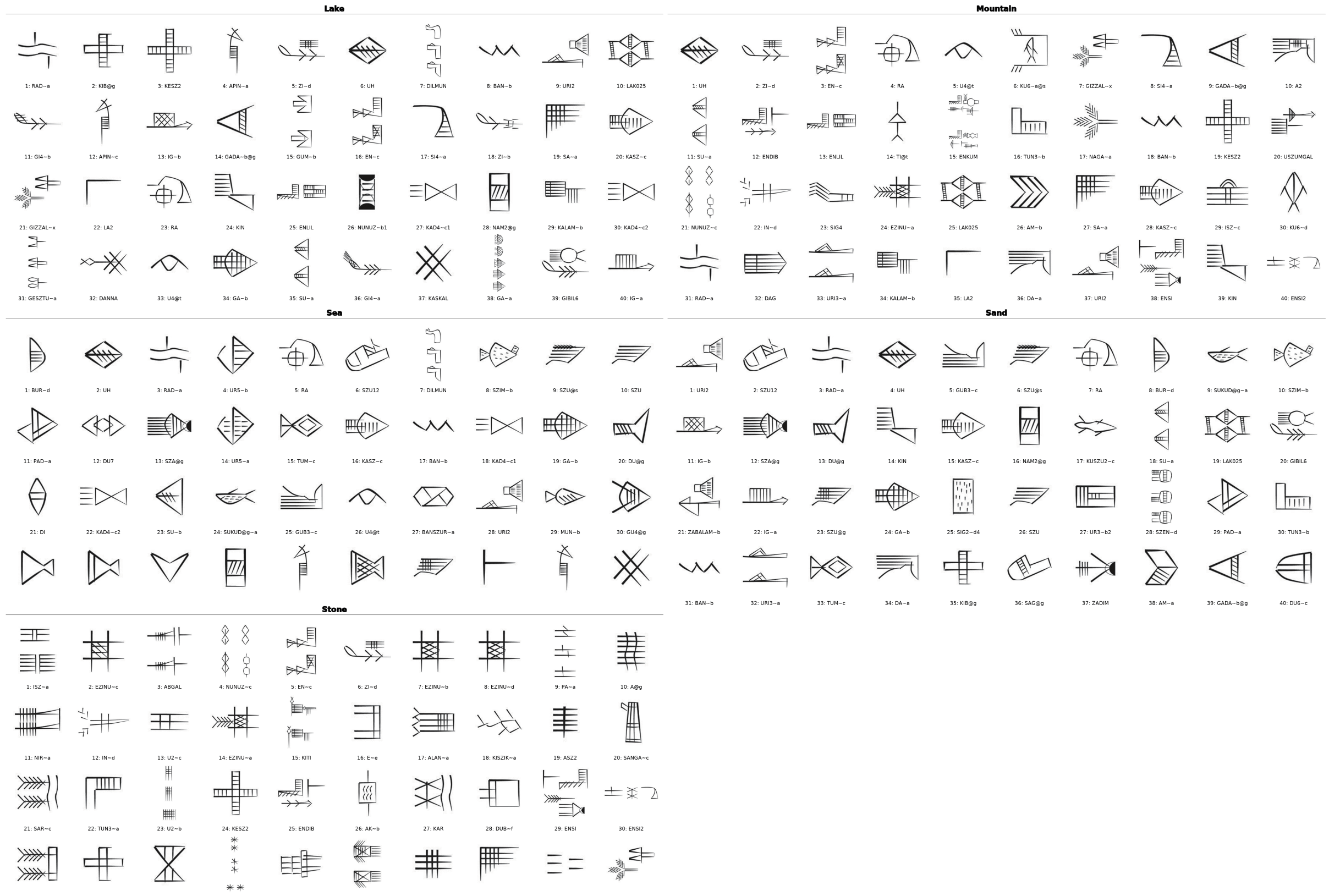

**Figure S8. Top 40 proto-cuneiform matches of scene categories guided by the generated sketches.** Sampled Proto-cuneiform pictographs for scene categories identified based on residualized category derived from the generated sketch embeddings. The number below each pictograph indicates its rank, reflecting the degree of similarity to the residualized embeddings (smaller, more similar). The original sign names are annotated beside each pictograph.

| | | Bird | | Dog | | Fish | | Flower | | Fruit | | Snake | | Mountain | | Stone | | Sand | | Lake | | Sea | | Total | |
|---|---|---|---|---|---|---|---|---|---|---|---|---|---|---|---|---|---|---|---|---|---|---|---|---|---|
| Strokes | Sketchability | Top1 | Top3 | Top1 | Top3 | Top1 | Top3 | Top1 | Top3 | Top1 | Top3 | Top1 | Top3 | Top1 | Top3 | Top1 | Top3 | Top1 | Top3 | Top1 | Top3 | Top1 | Top3 | Top1 | Top3 |
| 4 | 1 | 0.367 | 0.500 | 0.033 | 0.033 | 0.167 | 0.467 | 0.367 | 0.500 | 0.133 | 0.467 | 0.200 | 0.700 | 0.250 | 0.600 | 0.000 | 0.700 | 0.600 | 1.000 | 0.000 | 0.000 | 0.000 | 0.250 | 0.202 | 0.461 |
| | 2 | 0.233 | 0.433 | 0.067 | 0.133 | 0.133 | 0.267 | 0.433 | 0.533 | 0.000 | 0.367 | 0.300 | 0.733 | 0.233 | 0.667 | 0.115 | 0.654 | 0.667 | 0.967 | 0.000 | 0.067 | 0.000 | 0.100 | 0.199 | 0.445 |
| | 3 | 0.167 | 0.267 | 0.100 | 0.100 | 0.233 | 0.433 | 0.625 | 0.625 | 0.033 | 0.333 | 0.233 | 0.500 | 0.300 | 0.800 | 0.071 | 0.714 | 0.875 | 1.000 | 0.000 | 0.000 | 0.000 | 0.250 | 0.203 | 0.447 |
| 8 | 1 | 0.733 | 0.800 | 0.367 | 0.533 | 0.700 | 0.733 | 0.767 | 0.867 | 0.400 | 0.767 | 0.433 | 0.767 | 0.450 | 0.800 | 0.100 | 0.900 | 0.000 | 1.000 | 0.000 | 0.111 | 0.000 | 0.250 | 0.456 | 0.724 |
| | 2 | 0.633 | 0.900 | 0.267 | 0.467 | 0.467 | 0.667 | 0.067 | 0.333 | 0.100 | 0.800 | 0.233 | 0.567 | 0.433 | 0.800 | 0.154 | 0.731 | 0.500 | 0.967 | 0.000 | 0.133 | 0.067 | 0.333 | 0.337 | 0.666 |
| | 3 | 0.583 | 0.750 | 0.333 | 0.567 | 0.433 | 0.567 | 0.875 | 1.000 | 0.167 | 0.633 | 0.233 | 0.467 | 0.767 | 0.833 | 0.143 | 0.714 | 0.500 | 1.000 | 0.000 | 0.000 | 0.000 | 0.000 | 0.392 | 0.631 |
| 16 | 1 | 0.667 | 0.933 | 0.600 | 0.833 | 0.633 | 0.900 | 0.967 | 0.967 | 0.600 | 0.900 | 0.467 | 0.700 | 0.650 | 0.900 | 0.200 | 0.500 | 0.400 | 1.000 | 0.000 | 0.111 | 0.000 | 0.000 | 0.592 | 0.816 |
| | 2 | 0.733 | 0.767 | 0.567 | 0.900 | 0.533 | 0.767 | 0.967 | 1.000 | 0.400 | 0.867 | 0.333 | 0.500 | 0.700 | 0.900 | 0.115 | 0.577 | 0.433 | 0.900 | 0.000 | 0.233 | 0.233 | 0.400 | 0.460 | 0.712 |
| | 3 | 0.567 | 0.733 | 0.500 | 0.767 | 0.500 | 0.700 | 1.000 | 1.000 | 0.333 | 0.833 | 0.167 | 0.367 | 0.867 | 0.900 | 0.143 | 0.643 | 0.625 | 1.000 | 0.000 | 0.000 | 0.000 | 0.000 | 0.475 | 0.710 |

**Table S1. Top 1 and Top 3 decoding accuracy of generated sketches with semantic constraints across stroke counts and sketchability levels (related to Figure 3B).** The category-wise Top 1 and Top 3 accuracies of the generated sketches in matching their corresponding object or scene categories.

| | | Bird | | Dog | | Fish | | Flower | | Fruit | | Snake | | Mountain | | Stone | | Sand | | Lake | | Sea | | Total | |
|---|---|---|---|---|---|---|---|---|---|---|---|---|---|---|---|---|---|---|---|---|---|---|---|---|---|
| Strokes | Sketchability | Top1 | Top3 | Top1 | Top3 | Top1 | Top3 | Top1 | Top3 | Top1 | Top3 | Top1 | Top3 | Top1 | Top3 | Top1 | Top3 | Top1 | Top3 | Top1 | Top3 | Top1 | Top3 | Top1 | Top3 |
| 4 | 1 | 0.100 | 0.267 | 0.000 | 0.033 | 0.067 | 0.200 | 0.167 | 0.267 | 0.033 | 0.100 | 0.233 | 0.733 | 0.200 | 0.750 | 0.000 | 0.700 | 0.800 | 1.000 | 0.000 | 0.000 | 0.250 | 0.500 | 0.118 | 0.338 |
| | 2 | 0.067 | 0.233 | 0.000 | 0.000 | 0.000 | 0.367 | 0.000 | 0.167 | 0.000 | 0.333 | 0.233 | 0.633 | 0.100 | 0.600 | 0.077 | 0.462 | 0.833 | 1.000 | 0.000 | 0.133 | 0.000 | 0.167 | 0.120 | 0.371 |
| | 3 | 0.033 | 0.333 | 0.000 | 0.000 | 0.033 | 0.267 | 0.250 | 0.625 | 0.067 | 0.267 | 0.200 | 0.600 | 0.000 | 0.633 | 0.000 | 0.429 | 0.875 | 1.000 | 0.000 | 0.000 | 0.000 | 0.000 | 0.088 | 0.378 |
| 8 | 1 | 0.267 | 0.500 | 0.000 | 0.000 | 0.100 | 0.500 | 0.400 | 0.533 | 0.233 | 0.467 | 0.400 | 0.767 | 0.350 | 0.750 | 0.300 | 0.600 | 0.600 | 1.000 | 0.000 | 0.111 | 0.000 | 0.250 | 0.241 | 0.487 |
| | 2 | 0.267 | 0.467 | 0.000 | 0.000 | 0.167 | 0.400 | 0.000 | 0.167 | 0.300 | 0.567 | 0.300 | 0.533 | 0.067 | 0.433 | 0.000 | 0.538 | 0.500 | 0.967 | 0.000 | 0.033 | 0.000 | 0.167 | 0.175 | 0.417 |
| | 3 | 0.000 | 0.133 | 0.000 | 0.000 | 0.067 | 0.400 | 0.625 | 0.750 | 0.100 | 0.500 | 0.233 | 0.467 | 0.333 | 0.700 | 0.214 | 0.643 | 0.875 | 1.000 | 0.000 | 0.000 | 0.000 | 0.000 | 0.171 | 0.410 |
| 16 | 1 | 0.333 | 0.600 | 0.000 | 0.033 | 0.167 | 0.467 | 0.633 | 0.767 | 0.367 | 0.833 | 0.333 | 0.533 | 0.450 | 0.850 | 0.400 | 0.700 | 0.400 | 0.600 | 0.000 | 0.111 | 0.000 | 0.000 | 0.307 | 0.548 |

|  |  |  |  |  |  |  |  |  |  |  |  |  |  |  |  |  |  |  |  |  |  |  |  |  |
|---|---|---|---|---|---|---|---|---|---|---|---|---|---|---|---|---|---|---|---|---|---|---|---|---|
| 2 | 0.300 | 0.400 | 0.000 | 0.000 | 0.167 | 0.667 | 0.500 | 0.667 | 0.367 | 0.800 | 0.267 | 0.567 | 0.333 | 0.633 | 0.154 | 0.385 | 0.500 | 0.933 | 0.000 | 0.033 | 0.000 | 0.133 | 0.236 | 0.475 |
| 3 | 0.033 | 0.233 | 0.033 | 0.067 | 0.200 | 0.400 | 0.875 | 1.000 | 0.333 | 0.800 | 0.267 | 0.400 | 0.267 | 0.567 | 0.214 | 0.643 | 0.375 | 1.000 | 0.000 | 0.000 | 0.000 | 0.250 | 0.217 | 0.461 |

**Table S2. Top 1 and Top 3 decoding accuracy of generated sketches with perceptual constraints across stroke counts and sketchability levels (related to Figure 3B).** The category-wise Top 1 and Top 3 accuracies of the generated sketches in matching their corresponding object or scene categories.